\documentclass[twocolumn]{IEEEtran}

\usepackage[pdftex]{graphicx} % 插图命令 \includegraphics
\usepackage{amsmath} % Math Packages
\usepackage{algorithmic}
\usepackage{array} % Alignment Packages
\usepackage[caption=false,font=footnotesize]{subfig} % Subfigure Packages
\usepackage{dblfloatfix}
\usepackage{url} % simple URL typesetting

\usepackage{microtype}
\usepackage[utf8]{inputenc} % allow utf-8 input
\usepackage[T1]{fontenc}    % use 8-bit T1 fonts
\usepackage{paralist} % Compacted versions of enumerate and itemize.
\usepackage{marvosym} % 常用符号, 特别是打勾、打叉
\usepackage{bm} % 实现字体的斜体加粗
\usepackage{bbm} % fix \mathbb doesn't support digits
\usepackage{amssymb}   % 黑板粗体 \mathbb, 德文尖角体 \mathfrak, \backsim
\usepackage{mathrsfs} % 花体 \mathscr
\usepackage{tabularray} % tblr or talltblr or longtblr environment
\UseTblrLibrary{booktabs}       % professional-quality tables，也就是三线表
\usepackage{tabularx} % 自动计算列宽的表格包
\usepackage[flushleft]{threeparttable} % 给表格项添加注释
\usepackage{colortbl} % 表格行列颜色
\usepackage{makecell} % 表格内容换行
\usepackage{multirow} % 多行表格
\usepackage{nicefrac} % 斜着的分号
\usepackage[dvipsnames]{xcolor} % 字体颜色
\usepackage{soul} % highlight, strikeout, underline
\setul{0.5ex}{0.3ex}

\definecolor{citecolor}{RGB}{34,139,34}
\usepackage[pagebackref=true,breaklinks=true,letterpaper=true,colorlinks,
    citecolor=citecolor,bookmarks=false]{hyperref}
\usepackage{cleveref}  % for \cref
\crefformat{figure}{Fig.~#2#1#3}
\crefformat{equation}{Eq.~(#2#1#3)}
\crefformat{table}{Table~#2#1#3}
\crefformat{section}{Section~#2#1#3}
\crefformat{algorithm}{Algorithm~#2#1#3}
\crefformat{appendix}{Appendix #2#1#3}
\crefformat{subsection}{subsection~#2#1#3}
\usepackage{cellspace}
\usepackage[numbers,sort&compress]{natbib} % 代替 cite.sty

\usepackage{xspace} % 用于在 \newcommand 中添加空格

\usepackage{pifont} % http://ctan.org/pkg/pifont
\definecolor{Gray}{rgb}{0.9,0.9,0.9}
\definecolor{LightCyan}{rgb}{0.88,1,1}
\newcolumntype{a}{>{\columncolor{Gray}}c}
\newcolumntype{b}{>{\columncolor{white}}c}

\begin{document}

\setlength{\abovedisplayskip}{.5\baselineskip} % 调整公式与正文间的段前距离
\setlength{\belowdisplayskip}{.5\baselineskip} % 调整公式与正文间的段后距离

% \title{RepSLC: Re-Parameterized Sparse Local Contrast Networks for Spaceborne Dense Infrared Small Target Detection}
% \title{Scaling Down, Then Upscaling: Re-Parameterized Sparse Local Contrast Networks for Dense \\ Infrared Small Target Detection}
\title{Dual-Domain Perspective on Degradation-Aware Fusion: A VLM-Guided Robust Infrared and Visible Image Fusion Framework}
% \title{Zoom Out and In: Re-Parameterized Sparse Local Contrast Networks for Dense Infrared Small Target Detection}
% \title{Zooming Out and In: Re-Parameterized Sparse Local Contrast Networks for Dense Infrared Small Target Detection}
% \title{Zooming Image Out and In: Re-Parameterized Sparse Local Contrast Networks for Dense \\ Infrared Small Target Detection}

% \title{RepSLC: Re-Parameterized Sparse Local Contrast Networks for Spaceborne Infrared Small Target Detection}

% !TEX root = ../main.tex

\author{
  Tianpei~Zhang,
  Jufeng~Zhao,
  Yiming~Zhu,
  Guangmang~Cui

  % \thanks{
  %   This work was supported by
  %     % 国家自然科学基金
  %     the National Natural Science Foundation of China (), 
  %       % No. 62301036

  %   %
  %   and ``111'' Program B13022.
  %   \emph{The first two authors contributed equally to this work. (Corresponding author: Lang Wu)}
  %   }

  % % 724
  % \thanks{Yulei Qian is with Nanjing Marine Radar Institute, Nanjing, China
  % (e-mail: \href{mailto:yuleifly@126.com}{yuleifly@126.com}).
  % }

}

\maketitle

% !TEX root = ../main.tex

\begin{abstract}
Most existing infrared–visible image fusion (IVIF) methods assume high-quality inputs, and therefore struggle to handle dual-source degraded scenarios, typically requiring manual selection and sequential application of multiple pre-enhancement steps.  This decoupled pre-enhancement-to-fusion pipeline inevitably leads to error accumulation and performance degradation. To overcome these limitations, we propose \textbf{G}uided \textbf{D}ual-\textbf{D}omain Fusion (GD$^2$Fusion),  a novel framework that synergistically integrates vision-language models (VLMs) for degradation perception with dual-domain (frequency/spatial) joint optimization. Concretely, the designed Guided Frequency Modality-Specific Extraction (GFMSE) module performs frequency-domain degradation perception and suppression and discriminatively extracts fusion-relevant sub-band features. Meanwhile, the Guided Spatial Modality-Aggregated Fusion (GSMAF) module carries out cross-modal degradation filtering and adaptive multi-source feature aggregation in the spatial domain to enhance modality complementarity and structural consistency. Extensive qualitative and quantitative experiments demonstrate that GD$^2$Fusion achieves superior fusion performance compared with existing algorithms and strategies in dual-source degraded scenarios. The code will be publicly released after acceptance of this paper.
\end{abstract}

\vspace{-1\baselineskip}
% !TEX root = ../main.tex
% \bibliography{../reference.bib}

\section{Introduction}
% 由于成像机制的固有限制，单一模态的图像往往难以同时兼顾场景的全局感知与局部细节表达。例如，红外图像能够直接反映物体的热辐射特性，在弱光、遮挡等复杂环境下表现出较强的鲁棒性，但其在细节纹理和结构信息的呈现方面相对不足。相较之下，可见光图像能够提供丰富的纹理细节和视觉友好的色彩信息，然而其成像质量易受光照条件和遮挡因素的影响而显著下降。为克服单模态成像的局限性，红外与可见光图像融合（Infrared and Visible Image Fusion, IVIF）应运而生。该技术旨在充分挖掘并整合两种模态的互补优势，从而生成兼具红外鲁棒性与可见光细节性的高质量融合图像。更为重要的是，IVIF不仅提升了图像的可感知性和表达能力，还能够为智能驾驶、监控安防、目标检测与跟踪等下游计算机视觉任务提供更可靠的输入，从而显著增强其在真实场景中的应用性能。
In the domain of remote sensing detection, infrared and visible-light imaging constitute pivotal detection modalities. Due to the inherent limitations of imaging mechanisms, a single-modality image often fails to simultaneously capture both global contextual semantics information and fine local details. For instance, infrared images directly reflect the thermal radiation properties of objects, exhibiting strong robustness in challenging environments such as low illumination or occlusion. However, they are relatively insufficient in conveying detailed textures and structural information. In contrast, visible images provide rich texture details and human-perceptually compatible color information, yet their quality tends to degrade significantly under adverse illumination conditions or occlusion. To overcome the limitations of unimodal imaging, infrared and visible image fusion (IVIF) has emerged as a promising solution. This technique aims to fully exploit and integrate the complementary advantages of the two modalities, thereby producing high-quality fused images that simultaneously retain the robustness of infrared imaging and the fine details of visible imagery. More importantly, IVIF not only enhances image perceptibility and representational capacity, but also provides more reliable inputs for downstream computer vision tasks such as autonomous driving \cite{abrecht2024deep}, object detection \cite{hu2025datransnet,xiao2024background}, and tracking \cite{xiao2020review}, thus substantially improving their performance in real-world scenarios.

%在实际应用中，输入图像常不可避免地遭受多种质量退化——可见光图像在弱光条件下出现亮度不足或在强光下过度曝光，导致细节信息丢失；红外图像则易受传感器噪声、低对比度及环境干扰影响，使目标边缘模糊并降低前景—背景可分性。若将此类退化图像直接用于红外—可见光图像融合（IVIF），融合过程不仅难以体现两类模态的互补优势，反而可能同时保留或放大各自的退化模式并引入伪影，导致显著目标的丢失或语义扭曲，从而损害后续视觉任务的可靠性。以夜间场景为例：在救援场景中，可见光遥感因极低照度难以呈现地物边界，而受噪声影响的红外图像又会使遇险目标轮廓模糊——直接融合往往同时保留“结构缺失”与“目标轮廓虚化”的缺陷；在交通监控中，低光下可见光图像丢失车辆边缘 \cite{qu2024double} 而红外噪声又模糊行人轮廓，直接融合会引入伪影并降低如目标检测（mAP 等指标）等下游任务的性能 \cite{jiang2022degrade}。由此可见，设计一种能在多种复合退化条件下仍保持稳健表现的 IVIF 方法，不仅是实现鲁棒融合成像的核心挑战，也是推动该技术在复杂现实场景中可靠应用的关键。
In practical applications, input images are often unavoidably subjected to multiple forms of quality degradation. For instance, visible-light images may suffer from insufficient brightness under low-light conditions or over-exposure under different illumination conditions, leading to significant detail loss. Meanwhile, infrared images are prone to sensor noise, low contrast, and environmental interference, which blur object boundaries and weaken background separability. When such degraded images are directly used for IVIF, the fusion process not only struggles to preserve the complementary advantages of the two modalities but may also simultaneously retain or even amplify their respective degradation patterns, thereby introducing artifacts, losing salient targets, or distorting semantic content—ultimately impairing the reliability of downstream vision tasks. For instance, in nighttime remote sensing rescue scenarios, visible imagery fails to delineate object boundaries due to extremely low illumination, while noisy infrared imagery blurs the contours of distressed targets—direct fusion often retains both the “structural absence” and “blurred contour” defects. In traffic surveillance, low-light visible images lose vehicle edges \cite{qu2024double}, whereas noise-contaminated infrared images obscure pedestrian contours, and direct fusion introduces artifacts that degrade the performance of downstream object detection tasks \cite{jiang2022degrade}. These difficult scenarios underscore that designing an IVIF approach capable of maintaining robustness under diverse and compound degradations is not only a central challenge for robust fusion imaging but also a key to ensuring the reliable deployment of this technology in complex real-world scenarios.

\begin{figure}[!t]
\centering
\includegraphics[width = \columnwidth]{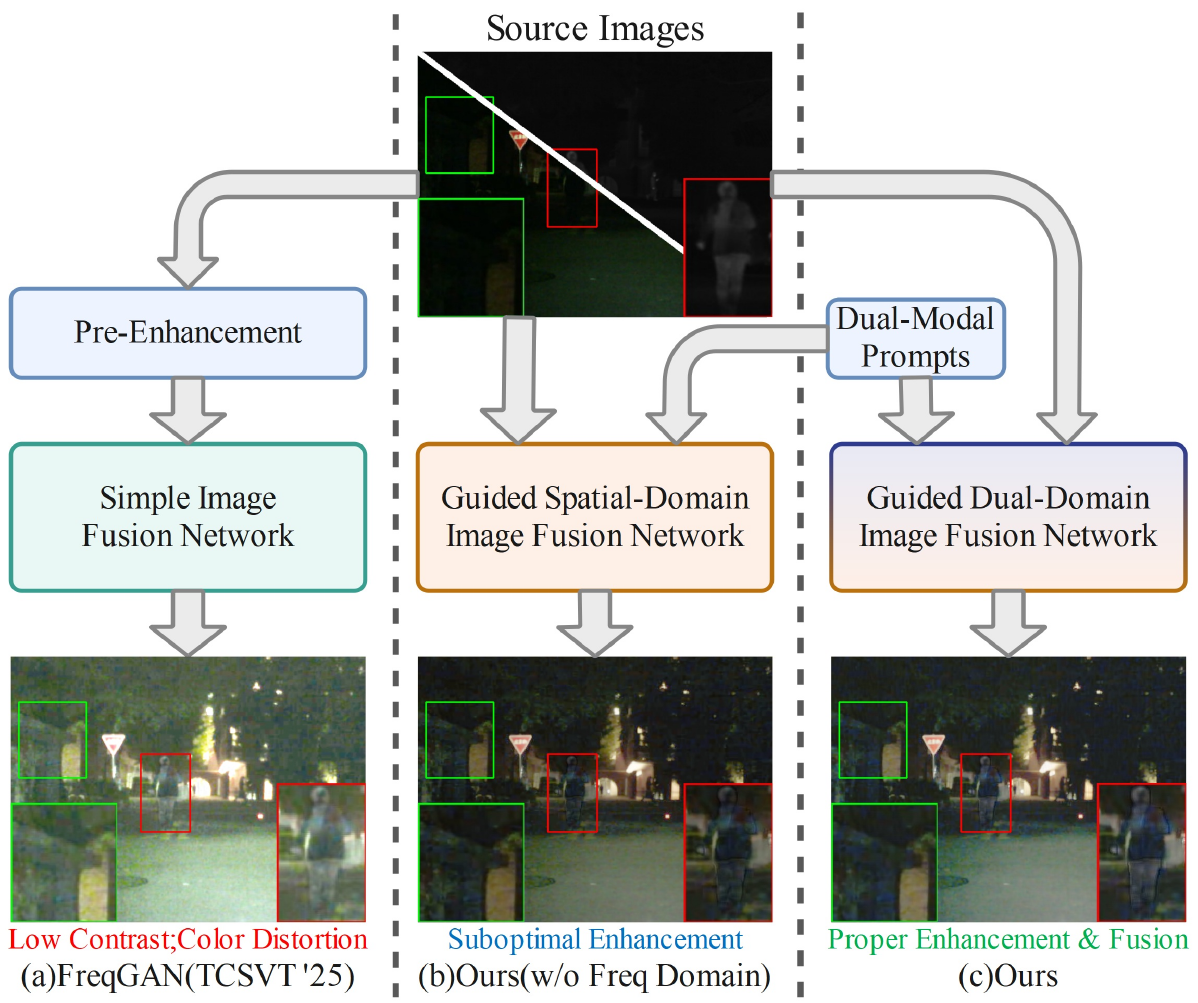}
%面对退化图像融合时的不同策略：（a）分离级联式的融合策略（b）我们所提出的基于VLM-Guided的双域分析的融合策略（c）基于VLM-Guided的仅有空间域分析的融合策略
\caption{Different strategies for degraded image fusion: (a) the conventional decoupled cascaded fusion paradigm; (b) our proposed VLM-guided dual-domain analysis fusion strategy; and (c) the VLM-guided fusion strategy relying solely on spatial-domain analysis.}
\label{fig1}
\end{figure}

%近年来，主流的深度学习红外—可见光图像融合（Infrared and Visible Image Fusion, IVIF）方法大体可归纳为四类：基于自编码器的重构式方法、基于生成对抗网络（GAN）的对抗式方法、以卷积神经网络（CNN）为主体的判别式方法，以及借助Transformer的全局建模方法。尽管这些方法在标准数据条件下取得了令人瞩目的表现，但大多隐含“高质量输入”假设：当源图像存在光照不足、过曝、对比度降低或传感器噪声等退化时，退化模式往往被直接传播到融合结果中，削弱了模态互补带来的增益。为解决源图像中的退化问题，这些隐含“高质量输入”假设的图像融合方法需要额外的预增强手段例如可见光的亮度增强、红外图像的去噪等先去除源图像退化再进行融合；同时也有工作面向单一源图像中的单类退化进行有针对性的处理,例如LVIF-Net仅针对IVIF中可能存在的可见光低光照问题；还有研究结合视觉-语言模型（VLM）实现文本引导的单一退化融合，如Text-IF解决源图像中存在的可见光低光，红外图像噪声等问题。然而，上述应对策略仍存在若干局限：
The current IVIF methods can be broadly categorized into four groups: autoencoder (AE)-based reconstruction methods \cite{li2021rfn,xu2022cufd}, generative adversarial network (GAN)-based adversarial methods \cite{ma2019fusiongan, ma2020ddcgan}, convolutional neural network (CNN)-based discriminative methods \cite{li2018densefuse, liu2023sgfusion, xu2020u2fusion}, and Transformer-based global modeling methods \cite{tang2022ydtr,tang2023datfuse,ma2022swinfusion}. Although these techniques achieve remarkable performance under standard benchmark conditions, most implicitly rely on a “high-quality input” assumption: when source images are afflicted by degradations such as low illumination, overexposure, reduced contrast, or sensor noise, these degradation patterns are often propagated into the fused outputs, thereby weakening the benefits of cross-modality complementarity. To address the degradation-induced performance drop of existing IVIF methods, fusion methods built upon the implicit assumption of “high-quality inputs” require additional pre-enhancement procedures—such as brightness enhancement for visible images \cite{Cui_2022_IAT} or denoising for infrared images \cite{liang2021swinir} to first remove degradations before performing the fusion process. However, there also exist studies that focus on addressing a single type of degradation within a single source image. For instance, LVIF-Net \cite{chang2024lvif} specifically targets the low-light condition that may occur in visible images during IVIF. Additionally, some works leverage vision–language models (VLMs) to enable text-guided fusion for handling individual degradations, such as Text-IF \cite{yi2024textif}, which addresses issues like low-light in visible images or noise in infrared images. However, the above remedies still suffer from several limitations:

%1.流程缺陷：遵循“图像预增强→图像融合”的分离级联范式，预处理与融合网络各自独立优化，易造成跨阶段的特征表征不一致，由此引发误差累积与性能退化；而面向特定退化的“一退化一运行”策略虽能在单一情形下取得针对性改进，但在实际应用中两模态常同时呈现多种或复合退化，仍然需要额外的人工预处理介入。
%2.泛化缺陷：预增强步骤依赖人工判别与参数调优，显著提高了实现复杂度，降低了系统稳定性与可复现性；同时缺乏从退化建模到融合决策的端到端一体化优化，使整体泛化能力受限。
%3.表征缺陷：多数方法仅在空间域进行特征提取与融合，未充分挖掘源图像中蕴含的频域信息，从而忽视了频域分析在退化抑制、细节保真与显著特征捕获方面的潜在增益。
%4.“流程缺陷—泛化缺陷—表征缺陷”相互作用，削弱了对退化特征的充分抑制与有效信息的高效传递，导致现有方法在双源多退化场景下表现不稳甚至失效，严重阻碍了红外–可见光图像融合在真实世界中的落地与应用。
\begin{enumerate}
    \item \textbf{Procedural Limitations:} Existing approaches commonly follow a decoupled cascade paradigm of “image pre-enhancement and fusion”, in which the pre-processing and fusion networks are independently optimized. This separation often leads to inconsistent feature representations across stages, resulting in error accumulation and performance degradation. Moreover, degradation-specific strategies that adopt a “one degradation, one execution” mechanism may yield improvements in isolated cases, yet in practical applications, both modalities frequently suffer from multiple or compound degradations simultaneously, thereby necessitating additional manual prep-rocessing.

    \item \textbf{Generalization Limitations:} The pre-enhancement step heavily relies on manual design and parameter tuning, which substantially increases implementation complexity while undermining system stability and reproducibility. Furthermore, the absence of an end-to-end joint optimization framework that directly integrates degradation modeling with fusion decision-making restricts the overall generalization capability.

    \item \textbf{Representation Limitations:} The majority of existing methods perform feature extraction and fusion solely in the spatial domain, without sufficiently leveraging the frequency-domain information embedded in the source images. Consequently, the potential benefits of frequency analysis—such as enhanced degradation suppression, detail fidelity, and salient feature preservation remain underexplored.

    \item \textbf{Joint Effect:} The interplay among these procedural, generalization, and representation limitations weakens the suppression of degradation features and the efficient transmission of useful information. As a result, current methods exhibit instability or even failure under multi-degradation scenarios across dual modalities, severely hindering the deployment and practical application of infrared–visible image fusion in real-world environments.
\end{enumerate}

%如 Fig.\ref{fig1} 所示，我们在退化场景下对典型融合策略进行对比分析。分离级联范式（Fig.\ref{fig1}(a)）虽可通过预增强在一定程度上抑制退化，但由于预增强与融合网络在目标函数与特征空间上彼此独立，易产生跨阶段表征不一致与误差传递；其表现为整体对比度不足与显著的色彩失真,甚至在融合结果中产生了过曝区域（红框的右边）。相较之下，仅依赖空间域建模的端到端融合框架（Fig.~\ref{fig1}(b)）避免了级联带来的分布漂移，获得了更自然的主观观感；然而由于忽略了与退化相关的频率维度信息，其在亮度恢复与显著目标信息保留方面仍显不足。基于上述观察，我们将提升退化图像融合的性能与便捷性的关键问题概括为两点： (1) 构建具备退化感知能力的端到端一体化框架，使退化抑制与特征融合在统一优化目标下协同完成，从根源上减少对繁琐预增强流程的依赖并减轻解耦导致的性能损失；(2) 引入频域—空间域的双视角互补建模与退化过滤，利用频域特征对纹理与结构信息进行选择性增强、对噪声与伪影进行抑制，以提升空间域表征的辨析力与稳健性，从而实现更全面且可靠的多模态融合。
As illustrated in Fig.~\ref{fig1}, we conduct a comparative analysis of typical fusion strategies under degraded scenarios. The cascaded paradigm (Fig.~\ref{fig1}(a)), though capable of partially suppressing degradation through pre-enhancement, suffers from inconsistency between the pre-enhancement and fusion networks in terms of objective functions and feature spaces, leading to cross-stage representation mismatch and error propagation. This manifests as insufficient global contrast and pronounced color distortion, and even results in overexposed regions in the fused image (highlighted on the right side of the red box). In contrast, the end-to-end fusion framework relying solely on spatial-domain modeling (Fig.~\ref{fig1}(b)) circumvents distribution shifts caused by cascaded processing and yields more natural subjective perception. Nevertheless, by neglecting frequency-domain cues closely associated with degradation, it still exhibits deficiencies in luminance restoration and salient target preservation. Based on these observations, we summarize the key issues for improving the performance and practicality of degraded image fusion into two aspects:
\textbf{Key Issue 1:}Developing an end-to-end unified framework with degradation-awareness, in which degradation suppression and feature fusion are jointly optimized under a common objective, thereby reducing dependence on laborious pre-enhancement procedures and alleviating the performance loss caused by decoupled processing;
\textbf{Key Issue 2:}Introducing dual-perspective complementary modeling and degradation filtering in the frequency–spatial domains, where frequency-domain representations selectively enhance texture and structural details while suppressing noise and artifacts, thus reinforcing the discriminability and robustness of spatial representations and ultimately achieving more comprehensive and reliable multimodal fusion.

%为应对上述挑战，本文提出了一种融合网络 Guided Dual-Domain Fusion (GD$^2$Fusion)。该框架利用VLMs强大的图文对齐能力获取提示特征，实现对源图像退化的感知与辅助抑制，从而规避了传统方法中依赖人工设计退化特征提取器的局限；并且GD$^2$Fusion 将频域分析与空间域建模有机结合——前者擅长解耦结构与细节成分，为有针对性的退化抑制与纹理恢复提供判别性特征，后者则侧重维护空间结构一致性与跨模态的语义对齐。具体来说，Guided Frequency Modality-Specific Extraction (GFMSE) 在模态特定提示的引导下，于频域视角建模单模态退化模式，并协同完成退化抑制与判别性特征提取；Guided Spatial Modality-Aggregated Fusion (GSMAF) 则在空间域视角中通过聚合模态提示感知并过滤跨模态特征的退化信息，同时自适应整合多源表示。围绕关键问题1（退化感知的一体化端到端建模），我们将网络组织为三条多层级、不同深度的特征路径：两条频域路径（基于 GFMSE）分别在浅至深的层次上实现单模态退化的逐级感知抑制与特征提取，另一条空间域路径（基于 GSMAF）则在对应深度实现跨模态退化的逐级过滤与特征融合，从而在统一优化目标下实现“退化抑制—特征融合”的协同推进。围绕关键问题2（频域—空间域的互补建模），我们将两条频域路径中增强后的频域特征注入到空间域融合路径，在多层级上实施跨域交互与互补增强：一方面利用频域成分提升空间域对细节纹理与结构的表达力，另一方面在各深度实现对噪声与伪影的高效过滤，最终获得在复杂退化条件下兼具鲁棒性与细节保真的融合结果。综上所述，我们工作的主要贡献如下：
To tackle these two key issues and bridge the gap between existing IVIF methods and real-world degraded scenarios, we proposed Guided Dual-Domain Fusion (GD$^2$Fusion). Within this framework, the powerful vision–language alignment capability of VLMs is exploited to obtain prompt features to enable perceptual awareness and auxiliary suppression of degradations in source images, thereby circumventing the limitations of traditional approaches that rely on manually designed degradation-specific feature extractors. Moreover, GD$^2$Fusion integrates frequency-domain analysis with spatial-domain modeling in a complementary manner: the former excels at decoupling structural and detail components, thereby providing discriminative features for targeted degradation suppression and texture restoration; while the latter emphasizes preserving spatial structural consistency and ensuring cross-modal semantic alignment. Specifically, the Guided Frequency Modality-Specific Extraction (GFMSE) module, under the guidance of modality-specific prompt features generated from VLMs, models modality-specific degradation patterns from a frequency-domain perspective, thereby jointly performing degradation suppression and discriminative feature extraction. In parallel, the Guided Spatial Modality-Aggregated Fusion (GSMAF) module operates from a spatial-domain perspective, where it leverages modality-aggregated prompts to filter degradation across modalities while adaptively integrating multi-source representations.
With respect to \textbf{Key Issue 1} (integrated end-to-end degradation-aware fusion modeling), GD$^2$Fusion is organized into three hierarchical and depth-varied feature pathways: two frequency-domain branches (based on GFMSE), which progressively perceive and suppress modality-specific degradations across shallow-to-deep layers while extracting representative features, and one spatial-domain branch (based on GSMAF), which progressively filters cross-modality degradations and performs feature fusion at corresponding depths. This design enables degradation suppression and feature fusion to be jointly optimized under a unified objective.
Regarding \textbf{Key Issue 2} (complementary frequency–spatial domain modeling), the enhanced frequency-domain features obtained from the two GFMSE pathways are injected into the spatial-domain fusion pathway to realize cross-domain interaction and complementary enhancement at multiple hierarchical levels. On the one hand, frequency components strengthen the spatial domain’s capacity to represent fine textures and structural details; on the other hand, multi-level integration ensures effective suppression of noise and artifacts. Consequently, the proposed framework yields fusion results that are simultaneously robust and detail-preserving under complex degradation conditions.
In summary, the primary contributions of this work are as follows:

%(1) 提出了名为GD^2Fusion的端到端融合框架。该框架借助基于视觉—语言模型（VLM）的文本提示实现对源图像退化的感知，并通过多层级的频域—空间域协同建模同时完成退化抑制与特征融合，从而在复杂退化条件下显著提升融合图像的表示能力与鲁棒性。
%(2) 设计了GFMSE模块。GFMSE 在频域视角下利用模态特定的提示信息对单模态退化进行建模与抑制，并在此基础上提取判别性频域特征，为后续跨模态融合提供高质量的频域表征。
%(3) 提出GSMAF模块。GSMAF 在空间域视角中通过聚合模态提示实现对跨模态退化信息的感知与过滤，并以自适应融合机制整合多源特征，从而增强模态互补性并保持结构一致性。
%(4) 在多个公开数据集上进行了充分的定性与定量对比实验。实验结果表明，本方法在面对多种退化场景时，不仅在视觉质量和细节保真度上优于若干代表性方法，而且在稳健性与下游可用性方面展现出显著改进。
\begin{itemize}
    \item We propose an end-to-end fusion framework termed GD$^2$Fusion, which leverages text prompts derived from Vision–Language Models (VLMs) to perceive degradations in source images. By performing multi-level joint modeling in both the frequency and spatial domains, GD$^2$Fusion simultaneously achieves degradation suppression and feature fusion, thereby substantially enhancing the representational capacity and robustness of fused images under complex degradation conditions.
    \item We design the GFMSE module. From the frequency-domain perspective, GFMSE employs modality-specific prompts to model and suppress single-modality degradations, while extracting discriminative frequency-domain features that provide high-quality representations for subsequent cross-modality fusion.
    \item We introduce the GSMAF module. From the spatial-domain perspective, GSMAF aggregates modality-aware prompts to perceive and filter cross-modality degradation information, and incorporates an adaptive fusion mechanism to integrate multi-source features, thus reinforcing modality complementarity while preserving structural consistency.
    \item We conduct extensive qualitative and quantitative experiments on multiple publicly available datasets. The results demonstrate that our method not only outperforms several representative approaches in terms of visual quality and detail fidelity under various degradation scenarios, but also exhibits superior robustness and enhanced usability for downstream tasks.
\end{itemize}
% !TEX root = ../main.tex
% \bibliography{../reference.bib}

\section{Related Works}
\subsection{Infrared and Visible Image Fusion}
\subsubsection{Image Fusion without Degradation Handling}
%深度学习的快速发展催生了大量基于学习的IVIF方法，这类方法在泛化能力与自动化程度上明显优于传统基于手工规则的融合策略。现有深度学习的 IVIF 方法大致可分为四类：基于自编码器（AE）、基于生成对抗网络（GAN）、基于卷积神经网络（CNN）以及基于 Transformer 的方法。
%• 基于自编码器（AE）的方法：该类方法通过编码器将输入图像映射到低维潜在表征，再由解码器重构融合图像。Liu等人提出了SGFusion利用显著性检测指导学习过程，实现了高质量的图像融合并减少了伪影；Wang等人提出了SwinFuse，其将自注意力机制引入特征提取阶段以增强融合图像的表示能力。
%• 基于生成对抗网络（GAN）的方法：GAN 通过生成器—判别器的对抗训练，鼓励融合结果在统计分布和视觉感知上更接近真实图像。生成器负责合成融合图像，判别器则判别合成图像与真实样本的差异；Ma等人提出FusionGAN，也是首次将生成对抗网络应用于图像融合领域；Liu等人提出的TarDAL通过以目标为核心的双对抗学习网络实现图像融合，增强了目标检测的准确性。GAN的训练不稳定性与模式崩溃等问题是需要权衡的挑战。
%• 基于卷积神经网络（CNN）的方法：CNN 通过局部感受野与逐层抽象提取多尺度的纹理与边缘特征，优势在于实现简洁、训练高效且易于工程化。Xu.等人建立了U2Fusion的融合框架，其实现了对多个任务的统一建模实现彼此互补，增强了融合质量；Tang等人提出的PIAFusion能够自适应地维持显著目标强度分布并保留背景中地纹理信息。
%• 基于 Transformer 的方法：Transformer 通过自注意力机制有效捕获长程依赖与跨区域的语义关系，因而在建模模态间全局互信息与上下文一致性方面具有天然优势。Tang等人提出了DATFuse，通过设计的双注意力残差模块捕获图像关键区域特征；Tang等人提出的YDTR通过Y型网络保留充足的细节信息。
%尽管上述方法在无明显退化的标准数据上取得了良好表现，但仍存在若干亟待解决的不足：（1）大多数工作侧重于在空间域内优化特征提取与融合策略，却未充分建模并利用频域信息，也忽视了频域与空间域之间的跨域互补性——因此在构建双域交互机制以提升信息表征能力方面仍有较大提升空间。（2）当前方法普遍隐含“高质量输入”假设，当源图像出现低光、过曝、低对比度或传感器噪声等退化情形时，融合结果常表现出退化残留或有效信息丢失，从而削弱了方法在复杂、真实场景下的泛化性与鲁棒性。

The rapid flourish of deep learning has spawned a large number of learning-based IVIF methods, which demonstrate markedly improved generalization and automation compared with traditional, hand-crafted fusion strategies \cite{wang2014fusion, selvaraj2020infrared}. Contemporary deep-learning IVIF approaches can be broadly grouped into four categories: autoencoder (AE)-based reconstruction methods, generative adversarial network (GAN)-based adversarial methods, convolutional neural network (CNN)-based approaches, and Transformer-based methods.
\begin{itemize}
    \item \textbf{AE-based methods} map input images into a low-dimensional latent representation via an encoder and reconstruct the fused image with a decoder. For example, Liu et al. propose SGFusion \cite{liu2023sgfusion}, which leverages saliency detection to guide the learning process and achieve high-quality fusion with reduced artifacts; Wang et al. introduce SwinFuse \cite{wang2022swinfuse}, which incorporates self-attention during feature extraction to strengthen the representational capacity of the fused output.
    \item \textbf{GAN-based methods} employ adversarial training between a generator and a discriminator to encourage fused outputs that better match the distribution and perceptual qualities of real images. Ma et al. introduce FusionGAN \cite{ma2019fusiongan}, the first application of GANs to image fusion, while Liu et al. present TarDAL \cite{liu2022target}, a target-centric dual-adversarial framework that enhances object detection accuracy in the fused results. Despite their advantages in texture preservation and naturalness, GANs bring challenges such as training instability and mode collapse that must be carefully managed.
    \item CNN-based approaches typically adopt end-to-end regression or reconstruction networks built from convolutional layers. Through local receptive fields and hierarchical abstraction, CNNs extract multi-scale texture and edge cues and guide multimodal integration using pixel-level losses, feature constraints, or attention mechanisms—advantages include implementation simplicity, efficient training, and engineering practicality. Xu et al. develop the U2Fusion framework \cite{xu2020u2fusion} to achieve unified modeling across multiple tasks and mutual complementarity, improving fusion quality; Tang et al. propose PIAFusion \cite{tang2022piafusion}, which adaptively preserves salient target intensity distributions while retaining background texture.
    \item Transformer-based methods leverage self-attention to capture long-range dependencies and cross-region semantics, offering natural strengths for modeling global cross-modality mutual information and contextual consistency. Tang et al. present DATFuse \cite{tang2023datfuse}, which captures key region features via a designed dual-attention residual module; they also propose YDTR \cite{tang2022ydtr}, a Y-shaped Transformer architecture that preserves abundant detail information.
\end{itemize}

Although the aforementioned methods have demonstrated promising performance on standard datasets without significant degradations, several critical limitations remain to be addressed: (1) Most approaches primarily focus on optimizing feature extraction and fusion strategies in the spatial domain, while insufficiently modeling and exploiting frequency-domain information, as well as overlooking the cross-domain complementarity between spatial and frequency representations. Consequently, there remains substantial room for improvement in establishing dual-domain interaction mechanisms to enhance representational capacity. (2) Existing methods generally operate under an implicit “high-quality input” assumption. When the source images suffer from degradations such as low illumination, overexposure, low contrast, or sensor noise, the fusion results often exhibit residual degradations or loss of salient information, thereby undermining the generalization ability and robustness of these methods in complex and real-world scenarios.

\subsubsection{Image Fusion under Degraded Scenarios}
%近年已有若干研究开始聚焦于退化图像融合问题，旨在缓解传统“预增强—融合”分离式流程在实际退化场景下导致的抑制不足与融合性能下降。针对这一问题的工作大致可分为两类：一类专注于单一退化的针对性处理，另一类则侧重于多种退化的泛化解决方案。
Recently, a number of studies have begun to focus on degraded-image fusion, aiming to mitigate the insufficient suppression and performance deterioration that arise from the conventional "pre-enhancement to fusion" decoupled pipeline in practical degraded scenarios. Efforts in this area can be broadly categorized into two classes: methods that target a single degradation type with specialized remediation, and approaches that seek generalized solutions capable of addressing multiple degradations.
\begin{itemize}
%面向单一退化的图像融合方法：该类方法通常针对某一明确的退化类型设计专门的网络结构或损失项，以实现对该类退化的定向校正与抑制。例如，Yang 等 \cite{yang2023unsupervised}提出通过引入自适应直方图均衡损失与联合梯度损失，以提升在可见光存在弱光退化情形下的融合性能；Tang 等人在 DIVFusion 中设计了基于 Retinex 理论的可见光照明分解子网络，旨在解决夜间低光环境下可见域信息稀缺问题；LVIF-Net 则采用图像分解网络来引导对融合图像照明分量的校正，从而改善亮度恢复与视觉一致性。总体来看，这类方法在其目标退化类型上能够取得针对性效果，但其专用性也限制了在复合或异质退化场景中的直接泛化能力。
\item Methods targeting single degradation types: these approaches typically design specialized network architectures or loss terms aimed at correcting and suppressing a specific, well-defined degradation. For example, Yang et al. \cite{yang2023unsupervised} introduce an adaptive histogram-equalization loss combined with a joint gradient loss to improve fusion performance under weak-illumination conditions in the visible domain; Tang et al., in DIVFusion \cite{tang2023divfusion}, design a Retinex-inspired visible-illumination decomposition sub-network to address information scarcity in nocturnal low-light scenarios; LVIF-Net \cite{chang2024lvif} employs an image-decomposition network to guide correction of the fused image’s illumination component, thereby enhancing brightness recovery and perceptual consistency. Overall, while these methods deliver effective, targeted improvements for their intended degradation type, their specialized nature limits direct generalization to composite or heterogeneous degradation scenarios.
%面向多种退化的图像融合方法：与面向单一退化的专用策略不同，此类方法倾向于通过更具泛化性的网络架构或训练范式，旨在应对多类或复合退化情形。具体而言，代表性工作往往借助具备强表达或生成能力的模型组件以实现对多样退化的鲁棒抵御：例如，DRMF 利用扩散模型的生成特性来补偿融合过程中出现的各类退化，从而实现多模态融合的稳健性；Text-IF 则首次将视觉—语言模型（VLM）引入退化图像融合，通过文本提示为网络提供高层语义先验，使其能在训练或推理阶段对多种退化类型进行感知与处理。这类方法通过提升模型的通用性与语义引导能力，在应对更广泛的现实退化场景方面展现出较强潜力。
\item Methods addressing multiple degradations: unlike specialized strategies designed for a single degradation type, these approaches favor more generalizable network architectures or training paradigms intended to cope with diverse or composite degradation scenarios. Representative works typically incorporate model components with strong expressive or generative capabilities to achieve robust resistance to varied degradations: for example, DRMF \cite{tang2024drmf} leverages the generative properties of diffusion models to compensate for a range of degradations arising during fusion, thereby improving multimodal fusion robustness; Text-IF \cite{yi2024textif} is among the first to introduce vision–language models (VLMs) into degraded-image fusion, using texts to supply high-level semantic priors that enable the network to perceive and address multiple degradation types during training or inference. By enhancing model generality and providing semantic guidance, these methods demonstrate substantial potential for handling a broader spectrum of real-world degradation conditions.
\end{itemize}

%尽管上述方法在一定程度上缓解了真实场景下的退化问题，但仍存在若干亟待解决的关键短板。（1）部分面向退化的融合方法（如 DIVFusion 或 Text-IF）在单次推理过程中通常仅能定向处理一种退化类型；当实际输入同时包含多类退化（例如可见光的低光与红外的噪声并存）时，这类方法仍需借助额外的预增强步骤以获得可接受的输入，从而无法实现从“退化感知—抑制”到“跨模态融合” 的端到端联合优化范式。（2）现有工作多数依赖单一的网络学习机制来弱化退化并增强有效信息，这在面对真实世界中多样且异构的退化类型时暴露出两类问题：一是可能需要为不同退化分别训练专用的预训练模型；二是在单一预训练模型中往往难以兼顾对多种退化的抑制能力，导致泛化性受限。这些问题的根源在于缺乏有效的语义级或模态自适应引导机制（例如基于 VLM 的提示），以便网络能可靠地区分退化特征与目标特征并在统一模型下实现对多类退化的协调抑制与平衡增强。
Although these methods partially mitigate degradation in practical scenarios, several critical shortcomings remain: (1) Some degradation-oriented fusion approaches (DIVFusion \cite{tang2023divfusion} or Text-IF \cite{yi2024textif}) typically address only a single degradation type per inference. When inputs exhibit concurrent, heterogeneous degradations (e.g., low-light in the visible channel together with noise in the infrared channel), such methods still rely on additional pre-enhancement steps to produce acceptable inputs, precluding an end-to-end joint optimization paradigm that unifies degradation perception–suppression with cross-modality fusion. (2) Most prior works depend on a single learning mechanism to both attenuate degradations and amplify informative signals, which in practice leads to two problems: (i) the potential need to train separate, degradation-specific pretrained models (such as DRMF \cite{tang2024drmf}); and (ii) the difficulty for a single pretrained model to robustly suppress multiple, diverse degradation types simultaneously, resulting in limited generalization. Fundamentally, these issues stem from the lack of effective semantic-level or modality-adaptive guidance mechanisms (for example, VLM-based prompts) that would enable a network to reliably discriminate degradation features from target features and to coordinate multi-type degradation suppression and balanced enhancement within a unified model.

\subsection{Wavelet-Transform in Image Fusion}
Since Mallat introduced the theory of multiresolution signal decomposition based on orthogonal wavelets \cite{mallat2002theory}, images can be robustly decoupled into low-frequency (approximation) and high-frequency (detail) subbands, thereby enabling the design of differentiated fusion rules in the frequency domain for structure and texture information. In one of the earliest works that brought wavelets into multisensor fusion, Li et al. \cite{li1995multisensor} used the discrete wavelet transform (DWT) to decompose multisensor images into scale–direction subbands and performed coefficient selection and weighting using rules such as maximum magnitude, energy, or regional consistency, followed by reconstruction to produce the fused image—thus establishing the classic “decompose–fuse–reconstruct” pipeline and inspiring numerous subsequent fusion criteria and evaluation methods. In recent years, wavelet-based techniques—owing to their controllability over luminance and texture and their interpretability—remain frequent baselines or decomposition modules within learnable frameworks \cite{luo2023infrared}. For example, MFIF-DWT-CNN \cite{avci2024mfif} couples DWT with convolutional neural networks in an end-to-end manner to leverage wavelets’ multiscale frequency representation together with CNNs’ nonlinear modeling, improving clarity and detail restoration for multi-focus fusion; WaveFusionNet \cite{liu2024wavefusionnet} integrates DWT with a multi-scale feature extraction network to enhance infrared–visible saliency and texture balance; W-Mamba \cite{zhang2025exploring} further combines wavelet decomposition with state-space modeling, designing heterogeneous feature-extraction and interaction mechanisms for different frequency components to achieve a better trade-off among texture preservation, structural consistency, and artifact suppression. 

Although IVIF approaches incorporating frequency-domain analysis—such as MFIF-DWT-CNN, WaveFusionNet, and W-Mamba—have demonstrated enhanced fusion performance under high-quality input conditions, they generally operate under the implicit assumption that the inputs are free from significant degradations, while paying limited attention to the potential of wavelet sub-bands for degradation suppression and informative feature enhancement. Therefore, in dual-source multi-degradation scenarios—for instance, when low illumination, overexposure, low contrast, or sensor noise coexist—the key unresolved challenge lies in fully exploiting the statistical properties of each sub-band within an end-to-end framework to model degradations, and in jointly optimizing frequency-domain processing with spatial-domain structural reconstruction and cross-modal consistency.

\subsection{Vision-Language Models}
%视觉—语言模型（Vision–Language Models, VLMs）通过在统一的表示空间中联合建模视觉与文本，实现了跨模态对齐与语义共享，从而推动了视觉问答\cite{antol2015vqa}、检索与开放词表识别等多模态任务的进展。早期的 VLM 多采用“双塔式”浅层结构与预定义视觉特征（如固定 CNN 提取器）来学习联合嵌入，侧重以对比/排序损失实现跨模态相似度度量。近年来兴起的 CLIP、BLIP、LLaVA 等模型在大规模图文对齐数据上进行预训练，显著强化了跨模态语义对齐与开放域泛化能力：CLIP 依赖对比式图文配对以学习判别性的全局对齐（Radford et al., 2021）\cite{radford2021clip}；BLIP 结合对比、匹配与文本生成目标以兼顾理解与生成（Li et al., 2022）\cite{li2022blip}；LLaVA 则以指令微调将视觉编码器与大语言模型连接，实现更强的指令理解与多轮交互（Liu et al., 2023）\cite{liu2023llava}。
%在此基础上，VLM 的高层语义先验与全局上下文建模逐步被引入低层视觉领域，用作“语义引导”的外部知识以辅助图像增强与图像融合：Cai等人\cite{cai2025degradation}等人通过VLM模型实现了退化感知的增强网络，能够进行针对性的退化识别及自适应增强；Shao 等人提出的 TFD$^2$Fusion 通过将视觉—语言模型的高级语义先验注入融合网络，旨在提升融合图像的语义丰富性与主观质量，但该工作并未针对退化图像融合问题利用 VLM 实现退化感知与抑制。Text-IF 则成为首个将 VLM 引入退化图像融合的尝试：其在特征融合阶段注入针对性退化的文本提示以指导网络去除特定类型的退化，然而该方法仅限于“单模态存在退化”的情形，且未将频域建模纳入其设计范式。总体来看，现有的 VLM 辅助 IVIF 研究在“双源多退化”场景下仍存在明显空白——主要表现在仅能处理单一退化、以及域建模较为片面（缺乏频域—空间域的协同）。因此，如何在双源多退化情形下实现 VLM 提示驱动的双域联合优化，既避免对退化特征的误强化，又保证语义引导与低层信号处理之间的一致性，从而产出“退化清晰且信息丰富”的融合图像，仍然是该方向亟待攻克的核心问题。
Vision–Language Models (VLMs) achieve cross-modality alignment and semantic sharing by jointly modeling visual and textual information within a unified representation space, thereby advancing a wide range of multimodal tasks such as visual question answering (VQA) \cite{antol2015vqa}. Early VLMs primarily adopted shallow “dual-tower” architectures with predefined visual features (e.g., fixed CNN extractors) to learn joint embeddings, relying on contrastive or ranking-based losses to measure cross-modality similarity. More recently, models such as CLIP \cite{radford2021clip}, BLIP \cite{li2022blip}, and LLaVA \cite{liu2023llava} have been pretrained on large-scale image–text alignment datasets, substantially strengthening cross-modality semantic alignment and open-domain generalization capabilities. For instance, CLIP leverages contrastive image–text pairs to learn discriminative global alignment; BLIP integrates contrastive, matching, and generative objectives to balance understanding and generation; and LLaVA employs instruction tuning to connect visual encoders with large language models, thereby enabling enhanced instruction following and multi-turn interactions.

Building upon these advances, the high-level semantic priors and global contextual modeling capabilities of VLMs have increasingly been introduced into low-level vision as external “semantic guidance” to assist tasks such as image enhancement and image fusion. For example, Cai et al. \cite{cai2025degradation} leveraged VLM-based priors to design a degradation-aware enhancement network capable of targeted degradation recognition and adaptive restoration; Shao \textit{et al.} proposed TFD$^2$Fusion \cite{shao2025infrared}, which enhances the semantic richness and perceptual quality of fused images by injecting high-level semantic priors from VLMs into the fusion network. However, this work does not exploit VLMs for degradation perception and suppression in degraded image fusion. Text-IF \cite{yi2024textif} represents the first attempt to incorporate VLMs into degraded image fusion: it introduces text prompts describing specific degradations during the feature fusion stage to guide the network in removing particular types of degradation. Nevertheless, this approach is restricted to scenarios where degradation exists in only one modality, and it does not incorporate frequency-domain modeling into its design paradigm. Overall, existing VLM-assisted IVIF studies exhibit a clear gap in addressing dual-source multi-degradation scenarios—primarily reflected in their ability to handle only single-type degradations and their one-sided domain modeling (lacking synergy between frequency and spatial domains). Consequently, achieving VLM-prompt-driven dual-domain joint optimization under multi-degradation conditions—while avoiding over-amplification of degradation features and maintaining consistency between semantic guidance and low-level signal processing—to generate fused images that are both degradation-free and information-rich, remains a key challenge to be addressed in this field.
%%%%%%%%%%%%%%%%%%%%%%%%%%%%%%%%%%%%%%%%%%
\begin{figure*}[!t]
\centering
\includegraphics[width=\textwidth]{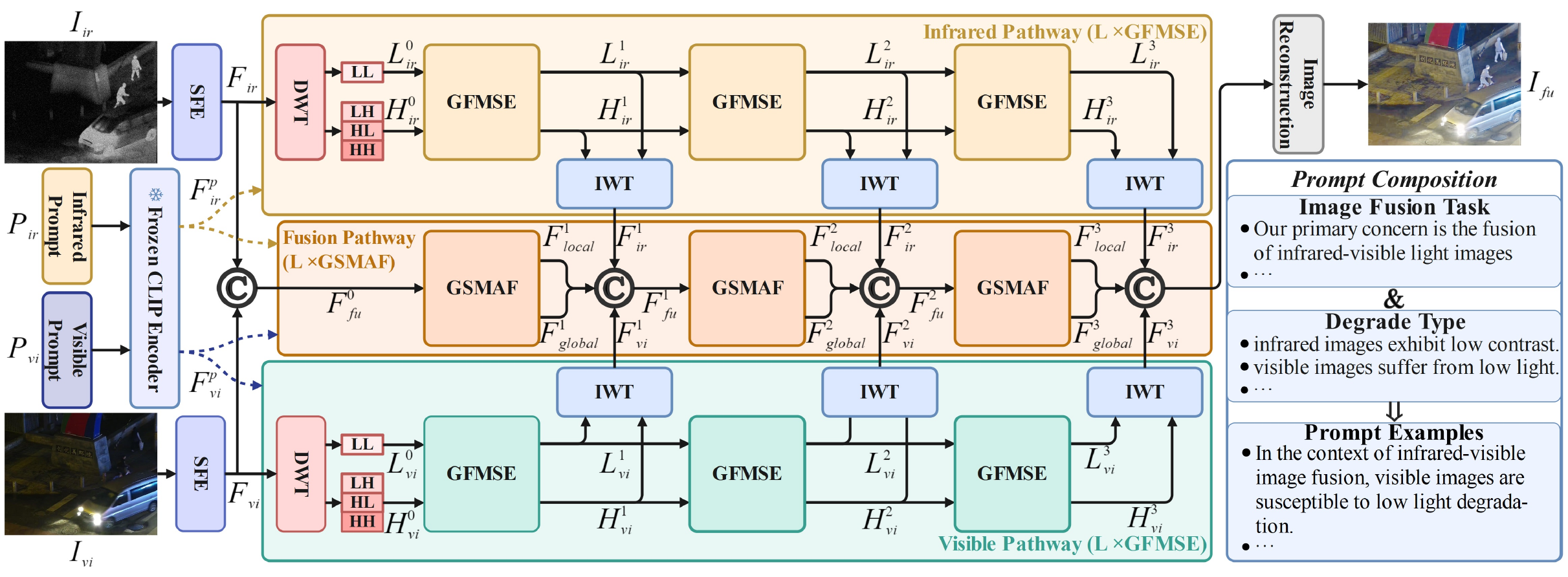}
%GD$^2$Fusion的网络结构，Prompt的组成在图的右下角展示，$L$表示堆叠的GFMSE/GSMAF的层数，在本文中，$L$被设置为3.
\caption{The architecture of GD$^2$Fusion, with the composition of the Prompt shown in the bottom-right corner of the figure. $L$ denotes the number of stacked GFMSE/GSMAF layers, which is set to 3 in this work.}
\label{network}
\end{figure*}

% !TEX root = ../main.tex
% \bibliography{../reference.bib}

\section{Method}
\subsection{Overall Architecture}
%如图~\ref{network} 所示，GD$^2$Fusion 给出了一个端到端的融合框架——这是首个将VLM 引导的退化感知与频域/空间域的双域联合优化在统一流水线上协同整合，以实现从退化识别、抑制到跨模态特征融合的全流程闭环工作。对于输入图片$I_{ir},I_{vi} \in B \times 3 \times H \times W$，其分别经过一个含有3 $\times$ 3的卷积操作的浅层特征提取模块（$SFE(\cdot)$）将输入图片投影至特征空间,其可以被表示为：
As illustrated in Fig.~\ref{network}, GD$^2$Fusion presents an end-to-end fusion framework—the first to synergistically integrate VLM-guided degradation perception with dual-domain (frequency/spatial) joint optimization within a unified pipeline, thereby enabling a complete closed-loop process from degradation identification and suppression to cross-modal feature fusion. Given the input images $I_{ir}, I_{vi} \in \mathbb{R}^{B \times 3 \times H \times W}$, each image is first processed by a shallow feature extraction module ($SFE(\cdot)$) consisting of a $3 \times 3$ convolution operation, which projects the inputs into the feature space. This process can be formulated as:
\begin{equation}
\label{eq:sfe}
F_{ir} = SFE(I_{ir}),F_{vi} = SFE(I_{vi})
\end{equation}
%其中$F_{ir},F_{vi} \in \in \mathbb{R}^{B \times C \times H \times W}$表示两种源图像的浅层特征表示。此外，对于搭配图像输入的提示输入$P_{ir},P_{vi} \in \in \mathbb{R}^{B \times w}$来说，他们经由冻结参数的具有强大文本图像对齐能力的CLIP编码器\cite{radford2021clip}进行编码,其过程可以被表示为：
where $F_{ir}, F_{vi} \in \mathbb{R}^{B \times C \times H \times W}$ denote the shallow feature representations of the two source images. In addition, the paired prompt inputs $P_{ir}, P_{vi} \in \mathbb{R}^{B \times w}$ are encoded through a CLIP encoder \cite{radford2021clip} with frozen parameters, which possesses strong text–image alignment capabilities. This process can be expressed as:
\begin{equation}
\label{eq:sfe}
F^{p}_{ir},F^{p}_{vi} = CLIP(P_{ir},P_{vi})
\end{equation}
%其中$F^{p}_{vi},F^{p}_{ir} \in \mathbb{R}^{B \times d}$表示用于退化指导的提示特征，这些提示特征被输入到每一个层级的GFMSE以及GSMAF模块中以进行不同层级下的退化感知指导，以实现充分的退化抑制。对于两种模态的特征提取路径Infrared Pathway以及Visible Pathway，两种模态的空间域浅层特征进行了对称的处理流程，为了简洁地表述对单模态浅层特征的处理流程，我们以红外浅层特征为例说明。首先红外浅层特征$F_{ir}$经由离散小波变换（DWT）实现频率分解，得到了表征亮度分布与结构轮廓的低频分量$LL^{0}_{ir} \in \mathbb{R}^{B \times C \times \frac{H}{2} \times \frac{W}{2}}$以及刻画纹理边缘的高频分量$LH^{0}_{ir},HL^{0}_{ir},HH^{0}_{ir} \in \mathbb{R}^{B \times C \times \frac{H}{2} \times \frac{W}{2}}$.随后低频分量及高频分量被输入进$L$级级联的GFMSE中与红外提示特征$F^{p}_{ir}$共同实现频域视角下由浅到深的退化感知抑制及频域信息提取，该流程可被表示为：
where $F^{p}_{vi}, F^{p}_{ir} \in \mathbb{R}^{B \times d}$ denote the prompt features utilized for degradation guidance. These prompt features are injected into each level of the GFMSE and GSMAF modules to provide degradation-aware guidance at multiple hierarchical stages, thereby facilitating effective degradation suppression. For the modality-specific feature extraction pathways, namely the Infrared Pathway and the Visible Pathway, the shallow spatial-domain features from both modalities undergo a symmetric processing scheme. To concisely illustrate the procedure for handling single-modality shallow features, we take the infrared shallow features as an example. Specifically, the infrared shallow features $F_{ir}$ are decomposed into the frequency domain via discrete wavelet transform (DWT) \cite{stankovic2003haar}, yielding the low-frequency component $LL_{ir} \in \mathbb{R}^{B \times C \times \frac{H}{2} \times \frac{W}{2}}$, which encodes luminance distribution and structural contours, as well as the high-frequency components $LH_{ir}, HL_{ir}, HH_{ir} \in \mathbb{R}^{B \times C \times \frac{H}{2} \times \frac{W}{2}}$, which capture texture and edge information. Subsequently, both the low- and high-frequency components are progressively processed through $L$ (set to 3 in this work) cascaded GFMSE modules, where they are jointly guided by the infrared prompt features $F^{p}_{ir}$ to achieve hierarchical degradation-aware suppression and frequency-domain feature extraction. This process can be formulated as:
\begin{equation}
\label{eq:model-specific pathway}
\begin{aligned}
&L^{0}_{ir} = LL_{ir}, H^{0}_{ir} = Concat(LH_{ir},HL_{ir},HH_{ir})\\
&L^{l}_{ir},H^{l}_{ir} = GFMSE(L^{l-1}_{ir},H^{l-1}_{ir},F^{p}_{ir})\\
&F^{l}_{ir} = IWT(L^{l}_{ir},H^{l}_{ir}),l \in [1,L]\\
\end{aligned}
\end{equation}
%其中$Concat(\cdot)$代表沿着batch维度的拼接操作，$L^{l}_{ir} \in \mathbb{R}^{B \times C \times \frac{H}{2} \times \frac{W}{2}}, H^{l}_{ir} \in \mathbb{R}^{3B \times C \times \frac{H}{2} \times \frac{W}{2}}$分别表示经由$l$层GFMSE处理后得到的低频分量和高频分量，$F^{l}_{ir} \in \mathbb{R}^{B \times C \times H \times W}$表示经由频域退化抑制及有效信息挖掘后通过小波反变换$IWT(\cdot)$得到的第$l$层的空间域红外特征.经过与infrared pathway相同的处理流程后，我们得到了对应的不同层级下的空间域可见光特征$F^{l}_{vi} \in \mathbb{R}^{B \times C \times H \times W}$.对于fusion pathway，两种模态的浅层特征$F_{ir},F_{vi}$通过沿通道维度的拼接得到浅层的融合特征$F^{0}_{fu} = Cat(F_{ir},F_{vi})$,其随后与两种模态的指导特征$F^{p}_{vi}, F^{p}_{ir}$被输入到$L$（在本文中被设置为3）级GSMAF中以实现聚合模态下对退化的综合感知过滤以及特征的多尺度融合，此外，同层级下的多尺度融合特征与经频域增强后的单模态特征组合为多源特征，以实现更充分的跨域交互，提升融合图像对结构/细节信息的表示能力，增强对残余退化信息的过滤性能，上述过程可以被表示为：
where $Concat(\cdot)$ denotes the concatenation operation along the batch dimension. $L^{l}_{ir} \in \mathbb{R}^{B \times C \times \frac{H}{2} \times \frac{W}{2}}$ and $H^{l}_{ir} \in \mathbb{R}^{3B \times C \times \frac{H}{2} \times \frac{W}{2}}$ represent the low-frequency and high-frequency components, respectively, obtained after processing through the $l$-th GFMSE. $F^{l}_{ir} \in \mathbb{R}^{B \times C \times H \times W}$ denotes the spatial-domain infrared feature via inverse wavelet transform $IWT(\cdot)$ at the $l$-th layer. Following the same processing pipeline as the infrared pathway, we obtain the corresponding spatial-domain visible features $F^{l}_{vi} \in \mathbb{R}^{B \times C \times H \times W}$ at different hierarchical levels.

For the fusion pathway, the shallow features of both modalities, $F_{ir}$ and $F_{vi}$, are concatenated along the channel dimension to form the initial fused representation $F^{0}_{fu} = Cat(F_{ir}, F_{vi})$. This representation, together with the modality-specific guidance features $F^{p}_{vi}$ and $F^{p}_{ir}$, is fed into the $L$-stage GSMAF to enable comprehensive degradation-aware filtering and multi-scale feature aggregation under a fused modality. Moreover, the multi-scale fused features at each level are further combined with the frequency-enhanced unimodal features to construct multi-source representations. This design facilitates more sufficient cross-domain interactions, enhances the fused image’s capacity to capture both structural and fine-grained details, and improves its ability to suppress residual degradations. The overall process can be formulated as:
\begin{equation}
\label{eq:fusion pathway}
\begin{aligned}
&F^{1}_{local},F^{1}_{global} = GSMAF(F^{0}_{fu},F^{p}_{ir},F^{p}_{vi})\\
&F' = Cat(F^{l-1}_{local},F^{l-1}_{global},F^{l-1}_{ir},F^{l-1}_{vi})\\
&F^{l}_{local},F^{l}_{global} = GSMAF(F',F^{p}_{ir},F^{p}_{vi}),l \in [2,L]\\
\end{aligned}
\end{equation}
%其中，$Cat(\cdot)$代表沿通道维度的拼接操作，$F^{l}_{local},F^{l}_{global} \in \mathbb{R}^{B \times C \times H \times W}$代表经$l$层GSMAF后输出的多尺度局部融合特征以及全局融合特征。最后，$F^{3}_{local},F^{3}_{global},F^{3}_{ir},F^{4}_{vi}$经过拼接后经由一个由三层连续的3*3卷积核+ReLU激活函数组成的图像重建模块得到最终的输出图像$I_{fu} \in \mathbb{R}^{B \times 3 \times H \times W}$,该过程可用公式表述为：
where $Cat(\cdot)$ denotes the concatenation operation along the channel dimension. The tensors $F^{l}_{local}, F^{l}_{global} \in \mathbb{R}^{B \times C \times H \times W}$ represent the multi-scale local fusion features and global fusion features obtained after the $l$-th GSMAF. Finally, $F^{L}_{local}, F^{L}_{global}, F^{L}_{ir}, F^{L}_{vi}$ are concatenated and passed through an image reconstruction module, which consists of three consecutive $3 \times 3$ convolutional layers followed by ReLU activations, to generate the final fused image $I_{fu} \in \mathbb{R}^{B \times 3 \times H \times W}$. This process can be formally expressed as:
\begin{equation}
\label{eq:image reconstruction}
\begin{aligned}
&I_{fu} = Re(Cat(F^{L}_{local},F^{L}_{global},F^{L}_{ir},F^{L}_{vi}))
\end{aligned}
\end{equation}
%其中$Re(\cdot)$表示图像重建模块。
where $Re(\cdot)$ denotes the image reconstruction module.

\subsection{Guided Frequency Modality-Specific Extraction}
\begin{figure}[!t]
\centering
\includegraphics[width = 0.8\columnwidth]{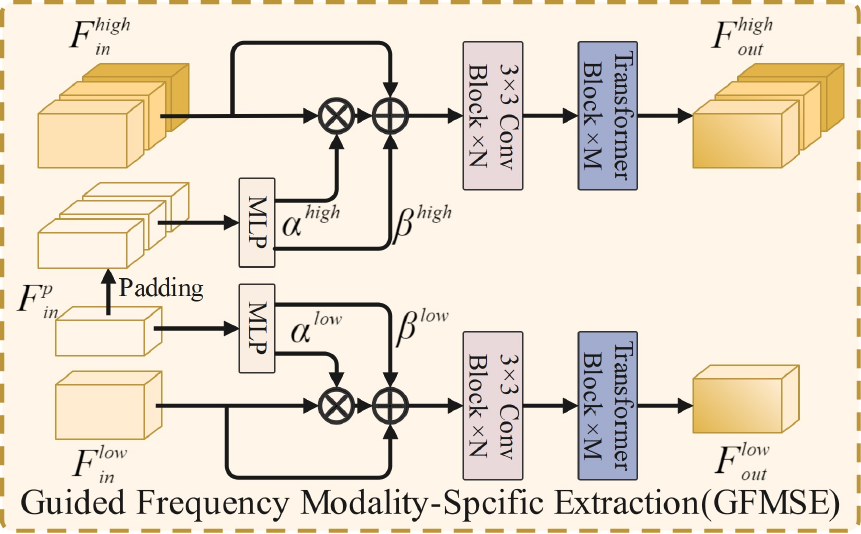}
%Guided Frequency Modality-Specific Extraction(GFMSE)的网络结构."Padding"指的是沿batch维度的膨胀操作,$N$和$M$分别表hi堆叠的卷积块和Transformer块的数量，在本文中被分别设置为3和2.
\caption{The network architecture of Guided Frequency Modality-Specific Extraction (GFMSE). "Padding" refers to the expansion operation along the batch dimension. $N$ and $M$ denote the numbers of stacked convolutional blocks and Transformer blocks, respectively, which are set to 3 and 2.}
\label{network_gfmse}
\end{figure}

%为充分利用不同频率分量的固有特性以同时提取低频的结构信息与高频的细节信息、抑制在各自频带中集中的退化成分（例如低频的对比度衰减与亮度偏移以及高频的传感器噪声\cite{he2025zero,gao2024research,zhang2006contrast}），并便于随后与空间域特征实现有效的域间互补以增强整体融合的鲁棒性与信息表征能力，我们设计了 Guided Frequency Modality-Specific Extraction（GFMSE）（结构见图~\ref{network_gfmse}）
%具体来说，输入指导特征$F^{p}_{in} \in \mathbb{R}^{B \times d}$通过一个MLP实现对低频图像特征$F^{low}_{in} \in \mathbb{R}^{B \times C \times \frac{H}{2} \times \frac{W}{2}}$进行指导的参数映射，这些参数随后通过仿射变换以实现有效的退化指导，而对于输入的高频图像特征$F^{high}_{in} \in \mathbb{R}^{3B \times C \times \frac{H}{2} \times \frac{W}{2}}$，我们通过对$F^{p}_{in}$进行batch尺度的膨胀以确保维度对齐并实现充分的退化指导，上述过程可以表示为：
To fully exploit the intrinsic characteristics of different frequency components—enabling the extraction of low-frequency structural information and high-frequency detail information, while suppressing degradation concentrated within their respective bands (e.g., contrast attenuation and luminance shifts in the low-frequency domain, as well as sensor noise in the high-frequency domain \cite{he2025zero,gao2024research,zhang2006contrast})—and to facilitate effective cross-domain complementarity with spatial-domain features for enhanced robustness and representational capacity of the overall fusion, we propose the Guided Frequency Modality-Specific Extraction (GFMSE) module (architecture shown in Fig.~\ref{network_gfmse}). Specifically, the input guidance features $F^{p}_{in} \in \mathbb{R}^{B \times d}$ are mapped to parameters through an MLP to guide the low-frequency image features $F^{low}_{in} \in \mathbb{R}^{B \times C \times \tfrac{H}{2} \times \tfrac{W}{2}}$. These parameters are subsequently applied via an affine transformation to achieve effective degradation-aware guidance. For the high-frequency image features $F^{high}_{in} \in \mathbb{R}^{3B \times C \times \tfrac{H}{2} \times \tfrac{W}{2}}$, we expand $F^{p}_{in}$ at the batch level to ensure dimensional alignment and thereby enable sufficient degradation-aware guidance. This process can be formally expressed as:
\begin{equation}
\label{eq:modality-specific guidance}
\begin{aligned}
 &\alpha^{low}, \beta^{low} = MLP(F^{p}_{in})\\
 &\alpha^{high}, \beta^{high} = MLP(Pad(F^{p}_{in}))\\
 &F^{low}_{guided} = \alpha^{low} \times F^{low}_{in} + \beta^{low} + F^{low}_{in} \\
 &F^{high}_{guided} = \alpha^{high} \times F^{high}_{in} + \beta^{high} + F^{high}_{in}\\
\end{aligned}
\end{equation}
%其中$Pad(\cdot)$表示沿batch维度的膨胀操作，低频仿射参数$\alpha^{low},\beta^{low}\in\mathbb{R}^{B\times C\times 1\times 1}$用于对低频分量施加仿射调制，以实现低频范围内的退化感知与抑制（例如红外对比度校正与可见光亮度补偿）并加强结构信息的表达；而高频仿射参数$\alpha^{high},\beta^{high}\in\mathbb{R}^{3B\times C\times 1\times 1}$ 则用于对高频分量施加条件化调制，以实现对高频聚集的退化（如红外噪声）的抑制及细节信息的增强。$F^{low}_{guided} \in \mathbb{R}^{B \times C \times H \times W},F^{high}_{guided} \in \mathbb{R}^{3B \times C \times H \times W}$分别表示经退化指导后的低频和高频图像特征。随后，两者经由N个连续的$3 \times 3$卷积块（其中包含串联的$3 \times 3$卷积和LeakyRelu激活函数）以及M个连续的Trasnformer块以实现不同尺度下的频域特征增强和退化抑制，以上过程可以被表示为：
where $Pad(\cdot)$ denotes the expansion operation along the batch dimension. The low-frequency affine parameters $\alpha^{low}, \beta^{low} \in \mathbb{R}^{B \times C \times 1 \times 1}$ are applied to the low-frequency components, facilitating degradation-aware suppression within the low-frequency range (e.g., infrared contrast correction and visible-light luminance compensation) while enhancing the representation of structural information. In contrast, the high-frequency affine parameters $\alpha^{high}, \beta^{high} \in \mathbb{R}^{3B \times C \times 1 \times 1}$ are imposed on the high-frequency components to achieve conditional modulation, thereby suppressing degradations concentrated in high-frequency bands (such as infrared noise) and reinforcing the preservation of fine details. The guided low- and high-frequency image features are represented as $F^{low}_{guided} \in \mathbb{R}^{B \times C \times \frac{H}{2} \times \frac{W}{2}}$ and $F^{high}_{guided} \in \mathbb{R}^{3B \times C \times \frac{H}{2} \times \frac{W}{2}}$, respectively. Subsequently, both are processed through $N$ consecutive $3 \times 3$ convolutional blocks (each consisting of a $3 \times 3$ convolution followed by a LeakyReLU activation) and $M$ successive Transformer blocks, enabling frequency-domain feature enhancement and degradation suppression across multiple scales. This overall process can be formally expressed as:
\begin{equation}
\label{eq:gfmse}
\begin{aligned}
 &F^{low}_{out} = TrmBlock^{M}(ConvBlcok^{N}_{3}(F^{low}_{guided}))\\
 &F^{high}_{out} = TrmBlock^{M}(ConvBlcok^{N}_{3}(F^{high}_{guided}))\\
\end{aligned}
\end{equation}
%其中$ConvBlock^{N}_{3}(\cdot)$表示N个连续的$3\times3$卷积块，类似的，$TrmBlock^{M}(\cdot)$表示M层连续的Transformer块，$F^{low}_{out} \in \mathbb{R}^{B \times C \times H \times W}$ and $F^{high}_{out} \in \mathbb{R}^{3B \times C \times H \times W}$分别表示GFMSE输出的高低频特征。
where $ConvBlock^{N}_{3}(\cdot)$ denotes $N$ consecutive $3 \times 3$ convolutional blocks, while $TrmBlock^{M}(\cdot)$ refers to $M$ successive Transformer layers. The outputs of GFMSE are represented as $F^{low}_{out} \in \mathbb{R}^{B \times C \times \frac{H}{2} \times \frac{W}{2}}$ and $F^{high}_{out} \in \mathbb{R}^{3B \times C \times \frac{H}{2} \times \frac{W}{2}}$, corresponding to the low- and high-frequency features, respectively.

\subsection{Guided Spatial Model-Aggregated Fusion}
\begin{figure}[!t]
\centering
\includegraphics[width = \columnwidth]{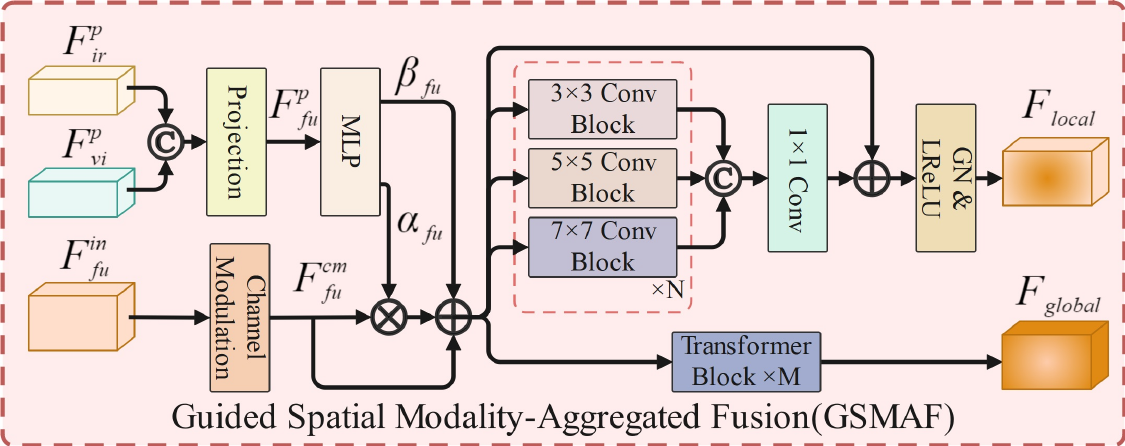}
%Guided Spatial Modality-Aggregated Fusion(GSMAF)的网络结构.
\caption{The network architecture of Guided Spatial Modality-Aggregated Fusion(GSMAF). $N$ and $M$ denote the numbers of stacked convolutional blocks and Transformer blocks, respectively, which are set to 3 and 2}
\label{network_gsmaf}
\end{figure}
%为在空间域实现对退化的感知与选择性过滤，并利用跨模态与跨域的互补性高效整合多源信息以提升融合图像的鲁棒性与表征能力，我们设计了结构如图~\ref{network_gsmaf} 所示的 Guided Spatial Modality-Aggregated Fusion (GSMAF) 模块。不同于传统的基于"特征相似度"的融合方法（例如DATFuse），GSMAF 在多尺度特征级别上评估由 VLM 语义提示所驱动的双模态特征的退化程度，并据此动态调节模态间特征的融合权重与过滤策略：在退化严重的区域，模块增强可信来源的响应并抑制退化成分；在信息互补性强的区域，则促进跨模态特征的协同聚合以增强融合特征的表征能力。具体来说，对于两种模态的输入指导特征$F^{p}_{vi},F^{p}_{ir} \in \mathbb{R}^{B \times d}$,我们首先完成跨模态的指导特征聚合得到聚合指导特征$F^{p}_{fu} \in \mathbb{R}^{B \times d}$，随后通过一个MLP实现非线性映射得到聚合模态的指导参数，输入的多源融合特征$F^{in}_{fu} \in \mathbb{R}^{B \times 4C \times H \times W}$通过由通道注意力和$1 \times 1$的卷积操作所组成的通道调制操作完成浅层的残余退化抑制及有效特征聚合，得到$F^{cm}_{fu} \in \mathbb{R}^{B \times C \times H \times W}$,其随后与聚合模态的指导参数完成仿射指导以实现对聚合模态下的退化程度的感知，上述过程可以用公式表示为：
To enable degradation-aware and selective filtering in the spatial domain, while efficiently integrating multi-source information through cross-modal and cross-domain complementarity to enhance the robustness and representational capacity of the fused image, we design the Guided Spatial Modality-Aggregated Fusion (GSMAF) module, as illustrated in Fig.~\ref{network_gsmaf}. Unlike conventional fusion approaches based on feature similarity (e.g., DATFuse \cite{tang2023datfuse}), GSMAF evaluates the degradation degree of bimodal features guided by VLM semantic prompts at the multi-scale feature level, and accordingly adapts the fusion weights and filtering strategies across modalities: in regions with severe degradation, the module strengthens the responses from reliable sources while suppressing degraded components; in regions with strong cross-modal complementarity, it promotes cooperative aggregation of features across modalities to enhance the representational power of the fused features. Specifically, for the input guidance features of the two modalities, $F^{p}_{vi}, F^{p}_{ir} \in \mathbb{R}^{B \times d}$, we first perform cross-modal aggregation to obtain the fused guidance feature $F^{p}_{fu} \in \mathbb{R}^{B \times d}$. This fused representation is then passed through an MLP to generate the nonlinear mapping that yields the guidance parameters of the aggregated modality. The multi-source fused feature input $F^{in}_{fu} \in \mathbb{R}^{B \times 4C \times H \times W}$ subsequently undergoes channel modulation, composed of channel attention and a $1 \times 1$ convolution, to achieve shallow residual degradation suppression and effective feature aggregation, resulting in $F^{cm}_{fu} \in \mathbb{R}^{B \times C \times H \times W}$. Finally, this representation is affine-guided with the aggregated modality guidance parameters to enable degradation-aware perception within the aggregated modality. The above process can be formally expressed as:

\begin{equation}
\label{eq:modality-aggregated guidance}
\begin{aligned}
 &F^{p}_{fu} = Proj(Cat(F^{p}_{ir},F^{p}_{vi}))\\
 &\alpha_{fu}, \beta_{fu} = MLP(F^{p}_{fu})\\
 &F^{cm}_{fu} = CM(F^{in}_{fu})\\
 &F^{guided}_{fu} = \alpha_{fu} \times F^{cm}_{fu} + \beta_{fu} + F^{cm}_{fu}\\
\end{aligned}
\end{equation}
%其中$Cat(\cdot)$表示沿通道维度的拼接操作，$Proj(\cdot)$表示通道投影操作，$\alpha_{fu},beta_{fu} \in \mathbb{R}^{B \times C \times 1 \times 1}$表示得到的聚合模态的指导参数,旨在自适应评估融合特征中的残余退化强度及增强互补特征。$F^{guided}_{fu} \in \mathbb{R}^{B \times C \times H \times W}$表示指导后的聚合模态特征。随后其分别经由由不同尺度的ConvBlock组成的局部特征聚合和由级联的TransformerBlock组成的全局特征聚合以得到不同尺度下的深层聚合特征$F_{local},F_{global} \in \mathbb{R}^{B \times C \times H \times W}$，上述过程可以被表示为：
where $Cat(\cdot)$ denotes the concatenation operation along the channel dimension, and $Proj(\cdot)$ represents the channel projection operation. $\alpha_{fu}, \beta_{fu} \in \mathbb{R}^{B \times C \times 1 \times 1}$ denote the guidance parameters of the aggregated modality, designed to adaptively assess the residual degradation intensity within the fused features and to enhance the complementary information, while $F^{guided}_{fu} \in \mathbb{R}^{B \times C \times H \times W}$ denotes the guided aggregated modality feature. Subsequently, $F^{guided}_{fu}$ is processed through local feature aggregation, implemented by ConvBlocks with multiple receptive-field scales, and global feature aggregation, realized via cascaded Transformer Blocks, to obtain the deep aggregated features at different scales, denoted as $F_{local}, F_{global} \in \mathbb{R}^{B \times C \times H \times W}$. The above process can be formulated as:
\begin{equation}
\label{eq:gsmaf}
\begin{aligned}
 &F'' = Cat(ConvBlock^N_{k}(F^{guided}_{fu}))\\
 &F_{local} = GN\&LReLU(Conv_1(F'') +F^{guided}_{fu})\\
 &F_{local} = TrmBlock^M(F^{guided}_{fu}),k=\{3,5,7\}\\
\end{aligned}
\end{equation}
%其中$ConvBlock^N_{k}(\cdot)$表示N个连续的卷积核尺寸为k的卷积块，为了获得不同感受野下的局部融合特征，我们在本文中将其设置为3，5，7总共三个分支。$Conv_1(\cdot)$表示$1\times1$的卷积操作，$GN\&LReLU(\cdot)$表示组归一化和LeakyReLU激活函数的组合.
where $ConvBlock^N_{k}(\cdot)$ denotes a sequence of $N$ consecutive convolutional blocks with kernel size $k$. To capture local fusion features under varying receptive fields, we employ three parallel branches with kernel sizes of 3, 5, and 7, respectively. $Conv_1(\cdot)$ represents a $1 \times 1$ convolution operation, while $GN\&LReLU(\cdot)$ refers to the combined operation of Group Normalization and the LeakyReLU activation function.

\subsection{Loss Function}
%为确保 GD$^2$Fusion 的输出既保留充分的强度信息、又保持丰富的纹理细节并避免色彩失真，我们将训练目标设计为由三项子损失加权组合而成：强度损失 $\mathcal{L}_{\text{int}}$、纹理损失 $\mathcal{L}_{\text{texture}}$ 与色彩损失 $\mathcal{L}_{\text{color}}$。具体而言，强度损失通过将融合图像与两源图像在每像素位置上的强度较大值作为参考来度量差异，从而鼓励融合结果继承源图像中的显著目标响应，其计算过程定义为：
To ensure that the output of GD$^2$Fusion preserves both sufficient intensity information and rich texture details while avoiding color distortion, the training objective is formulated as a weighted combination of three sub-losses: intensity loss $L_{int}$, texture loss $L_{text}$, and color loss $L_{color}$. Specifically, the intensity loss measures the discrepancy between the fused image and the per-pixel maximum intensity of the two source images, thereby encouraging the fusion result to inherit prominent target responses from the source images. Its computation is defined as follows:
\begin{equation}
\label{eq:Int loss}
 L_{int} = \frac{1}{HW}||I_{fu}-max(I^{ref}_{ir},I^{ref}_{vi})||_{1}
\end{equation}
%其中，$H$ 与 $W$ 分别表示图像的高度与宽度，$I_{fu}, I^{\text{ref}}_{ir}, I^{\text{ref}}_{vi} \in \mathbb{R}^{B \times 3 \times H \times W}$ 分别表示融合图像以及无退化的红外与可见光参考图像，$\max(\cdot)$ 表示像素级的取最大值操作。随后，在纹理损失 $\mathcal{L}_{\text{texture}}$ 中，我们通过计算融合图像与源图像在梯度域的逐像素最大值之间的差异，以充分保留源图像中的纹理细节信息，其具体公式定义为：
where $H$ and $W$ denote the height and width of the image, respectively, and $I_{fu}, I^{ref}_{ir}, I^{ref}_{vi} \in \mathbb{R}^{B \times 3 \times H \times W}$ represent the fused image and the degradation-free infrared and visible reference images. The operator $\max(\cdot)$ indicates the per-pixel maximum selection. Subsequently, in the texture loss $L_{text}$, the discrepancy between the per-pixel maximum of the gradients of the fused image and the source images is computed to effectively preserve the texture details of the source images. The specific formulation is defined as follows:
\begin{equation}
\label{eq:Texture loss}
 L_{text} = \frac{1}{HW}||\nabla I_{fu}-max(\nabla I^{ref}_{ir},\nabla I^{ref}_{vi})||_{1}
\end{equation}
%其中 $\nabla$ 表示 Sobel 梯度算子。值得注意的是，现有许多彩色图像融合管线常将可见光图像从 RGB 空间转换到 YCbCr 空间，并仅以代表亮度的 Y 分量作为网络输入进行融合；但该做法对色度分量缺乏约束，容易在预处理与融合过程中产生色彩偏移，并与预处理引入的颜色误差叠加，最终导致融合结果出现显著的色彩失真，从而降低主观视觉质量。为在退化图像融合情形下抑制上述色彩失真并提高色彩保真度，我们在 YCbCr 空间对色度通道施加显式约束：令 $(\cdot)_{CbCr}$ 表示将图像转换到 YCbCr 后取色度分量（Cb、Cr），则颜色损失可定义为色度通道的逐像素差异的 $L_1$ 距离：
where $\nabla$ denotes the Sobel gradient operator. It is noteworthy that many existing color image fusion pipelines convert the visible image from the RGB space to the YCbCr space and perform fusion using only the luminance (Y) channel as network input. However, this approach imposes no constraints on the chrominance channels, which can lead to color shifts during fusion; such shifts may accumulate with preprocessing-induced color errors, ultimately causing noticeable color distortion in the fused results and degrading subjective visual quality. To mitigate these color distortions and enhance color fidelity in degraded image fusion scenarios, we impose explicit constraints on the chrominance channels in the YCbCr space. Let $(\cdot)_{CbCr}$ denote extracting the chrominance components (Cb, Cr) after conversion to YCbCr; then, the color loss can be defined as the per-pixel $L_1$ distance in the chrominance channels:
\begin{equation}
\label{eq:Color loss}
L_{color}=\frac{1}{HW}||(I_{fu})_{CbCr}-(I^{ref}_{vi})_{CbCr}||_{1}
\end{equation}

%总的来说，GD$^2$Fusion的总损失函数可以被如下表示为：
In summary, the overall loss function of GD$^2$Fusion can be formulated as follows:
\begin{equation}
\label{eq:Loss Function}
 L = \gamma L_{int} + \lambda L_{text} + \theta L_{color}
\end{equation}
%其中$\gamma,\lambda,\theta$分别表示三种子损失函数的权重。
where $\gamma$, $\lambda$, and $\theta$ denote the respective weights of the three sub-loss functions.
% !TEX root = ../main.tex
% \bibliography{../reference.bib}

\section{Experiment}
\subsection{Experimental Setup and Datasets}
%为系统评估 GD$^2$Fusion，我们分别训练了**退化感知**的融合模型与**高质量输入假设**下的融合模型。对于前者，从 EMS 数据集 \cite{textif} 随机选取 2,278 张图像构成训练集，样本为配准的源图像对，其中可见光包含低光与过曝两类退化，红外包含低对比度与噪声两类退化；对于后者，从 LLVIP 数据集中随机选取 11,025 张图像进行训练。训练过程中，将输入裁剪为 $96\times96$ 的图像补丁送入网络，批大小设为 16；两类模型的训练轮数分别为 200（退化感知）与 100（高质量输入假设）。学习率固定为 $2.5\times10^{-4}$；损失函数中的超参数 $\gamma,\lambda,\theta$ 分别设为 5、6、5。所有实验均在单张 NVIDIA GeForce RTX 4090D GPU 上完成。 为验证退化感知模型在退化场景下的有效性，我们从 EMS 数据集中分别随机抽取 69、69、35、35 对图像，对应四种组合情形：可见光低光 & 红外低对比度、可见光低光 & 红外噪声、可见光过曝 & 红外低对比度、以及可见光过曝 & 红外噪声。对于基于高质量输入假设的模型，为展示其性能与泛化性，我们在 MSRS 与 LLVIP 数据集上分别随机抽取 45 与 50 张图像进行对比评测。
To systematically evaluate GD$^2$Fusion, we trained two distinct models: a degradation-aware fusion model and a high-quality input assumption fusion model. For the former, we randomly selected 2,278 registered source image pairs from the EMS dataset \cite{yi2024textif} as the training set, where the visible images exhibit two types of degradations (low-light and overexposure) and the infrared images contain two types of degradations (low contrast and noise). For the latter, 11,025 images were randomly sampled from the LLVIP dataset \cite{jia2021llvip} for training. During training, inputs were cropped into $96\times96$ patches and fed into the network with a batch size of 16. The two models were trained for 200 epochs (degradation-aware) and 100 epochs (high-quality input assumption), respectively. The learning rate was fixed at $2.5\times10^{-4}$, and the loss function hyperparameters $\gamma$, $\lambda$, and $\theta$ were set to 5, 5, and 6, respectively. All experiments were conducted on a single NVIDIA GeForce RTX 4090D GPU.

To verify the effectiveness of the degradation-aware model under degraded scenarios, we randomly selected 69, 69, 35, and 35 image pairs from the EMS dataset, corresponding to four degradation combinations: visible low-light \& infrared low-contrast, visible low-light \& infrared noise, visible overexposure \& infrared low-contrast, and visible overexposure \& infrared noise. For the high-quality input assumption model, to demonstrate its performance and generalization ability, we randomly selected 45 images from the MSRS dataset \cite{tang2022piafusion} and 50 images from the LLVIP dataset \cite{jia2021llvip} for comparative evaluation.

\subsection{Comparative Methods and Evaluation Metrics}

\begin{table*}[!t]
  \setlength{\abovecaptionskip}{0.1cm}  
  \renewcommand\arraystretch{1.2}
  \footnotesize
  \centering
  \vspace{-1\baselineskip}
\caption{Configurations for all compatative experiments}
\vspace{-1\baselineskip}
\label{tab:comparative methods}
\tabcolsep=0.3cm
\renewcommand\arraystretch{1.2}
\begin{center}
\resizebox{\textwidth}{!}{
\begin{tabular}{ll}
\toprule[1pt]
Methods(\textit{Source \& Year \& Type})         & Key parameters configurations   \\ \hline
GANMcC \cite{ma2020ganmcc} \textit{TIM'2020, GAN-based} & Patch Size: 120 $\times$ 120, $b = 32, p = \frac{1}{2}, M = 10, \gamma = 100, \beta_{1} = 1, \beta_{2} = 5, \beta_{3} = 4, \beta_{4} = 0.3$ \\ 
SDNet \cite{zhang2021sdnet} \textit{IJCV'2021, CNN-based}     &  Patch size: \{60 $\times$ 60, 120 $\times$ 120\}, $\alpha = [0.5,0.5,0.5,1,1], \beta = [10,80,50,3]$\\ 
CSF \cite{xu2021classification} \textit{TCI'2021, AE-based} & Patch Size: 128 $\times$ 128, $\lambda = 25, \sigma = 0.0001, L = 30, \alpha=0.7$ \\ 
DATFuse \cite{tang2023datfuse} \textit{TCSVT'2023, Transformer-based} & Patch Size: 120 $\times$ 120, $C = 16, R = 4, \alpha=1, \lambda = 100, \gamma = 10$ \\ 
ITFuse \cite{tang2024itfuse} \textit{PR'2024, Transformer-based}     &   Patch size: 120 $\times$ 120, $ N = 3, C = 16, R = 4, L = 3, \epsilon = 0.001, \alpha = 1, \beta = 1, \gamma = 4, \lambda = 4$ \\
Text-IF \cite{yi2024textif} \textit{CVPR'2024, Transformer-based}     &   Patch size: 96 $\times$ 96, $\alpha_{int} = \{8,4,8,6\}, \alpha_{SSIM} = \{1,0,1,1\}, \alpha_{grad} = \{10,2,10,10\}, \alpha_{color} = \{12,12,12,12\}$               \\
FreqGAN \cite{10680110} \textit{TCSVT'2025, GAN-based}     &   Patch size: 256 $\times$ 256, $\lambda_{1} = 10, \lambda_{2} = 16, \lambda_{3} = 14, r = 0.4$\\
\hline
\rowcolor[rgb]{0.9,0.9,0.9}$\star$ \textbf{GD$^2$Fusion(Ours), Transformer-based} & Patch size: 96 $\times$ 96, $L=3, \gamma = 5, \lambda = 5, \theta = 6$\\
\bottomrule[1pt]
\end{tabular}}
\end{center}
\end{table*}

\begin{figure*}[!t]
\centering
\includegraphics[width=0.75\textwidth]{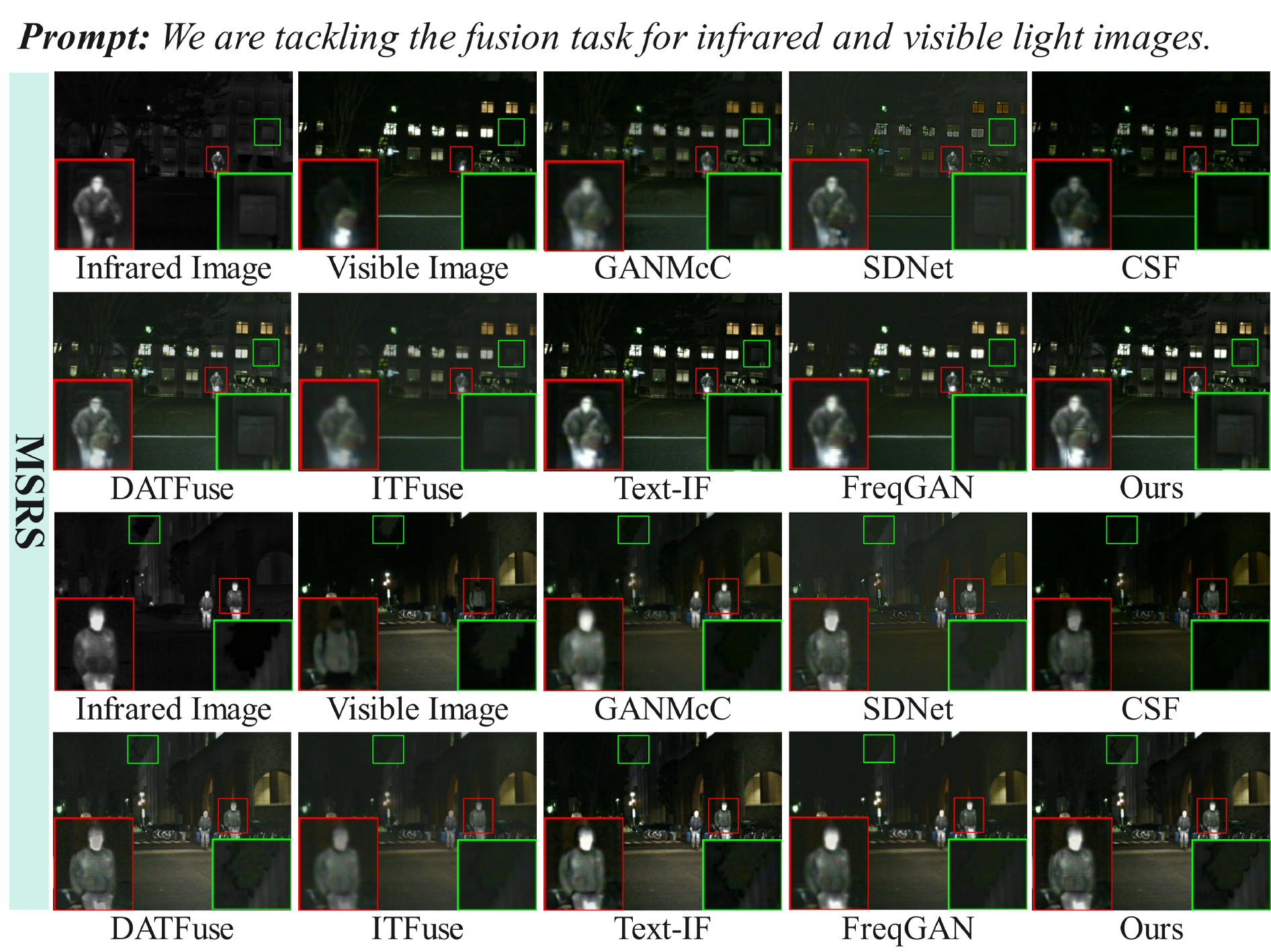}
%图中展示了 GD$^2$Fusion 与其他七种比较图像融合方法在 MSRS 数据集上进行融合实验的定性对比结果，实验场景基于高质量图像输入的假设设定。图中所示的 Prompt 表示输入到我们网络中的文本提示信息，用于指导融合过程。为了突出关键差异，我们在融合结果中对若干感兴趣区域采用红色与绿色矩形框进行标注，并对其进行了局部放大，从而便于在细节层面开展直观的视觉比较与分析。
\caption{Qualitative comparison results of GD$^2$Fusion against seven comparative image fusion methods on the MSRS dataset, under the assumption of high-quality input images. The Prompt shown in the figure denotes the textual guidance fed into our network to guide the fusion process. To highlight key differences, several regions of interest are annotated with red and green bounding boxes in the fused results, and subsequently enlarged to facilitate intuitive visual inspection and detailed analysis.}
\label{qualitative_msrs}
\end{figure*}

%我们在表 \ref{tab:comparative methods} 中列示了用于性能对比的融合方法，覆盖两类代表性方案：其一为基于高质量输入假设的先进融合网络，其二为面向单一退化类型设计的融合模型。对于退化感知的融合模型，由于缺乏可用作参考的“完全去除退化的理想真值图像”，难以采用有参考指标进行评估。为在退化场景下客观衡量各方法的优劣，我们选取四种常用的无参考评价指标开展定量对比：平均梯度 $AG$、边缘强度 $EI$、标准差 $SD$ 与空间频率 $SF$。对于高质量输入假设的融合模型，我们采用信息论指标——特征互信息 $FMI$ 与互信息 $MI$；基于图像特征的指标——梯度相似性度量 $Q_{abf}$；两类结构相似性指标——Piella 指标 $Q_{W}$ 与 Yang 指标 $Q_{y}$；以及基于人类视觉感知的指标——视觉信息保真度 $VIF$，以实现对融合图像质量的较为全面的量化评估。上述十项指标均遵循“数值越大表示性能越优”的判定准则。
The fusion methods employed for performance comparison are summarized in Tab.~\ref{tab:comparative methods}, encompassing two representative categories: (1) advanced fusion networks developed under the assumption of high-quality inputs, and (2) models specifically designed to address a single type of degradation. For the degradation-aware fusion models, due to the absence of “ideal ground-truth images with completely removed degradations” as references, reference-based metrics cannot be applied. To objectively evaluate the relative merits of different approaches under degraded conditions, we adopt four widely used no-reference metrics: average gradient ($AG$ \cite{cui2015detail}), edge intensity ($EI$ \cite{rajalingam2018hybrid}), standard division ($SD$ \cite{rao1997SD}), and spatial frequency ($SF$ \cite{eskicioglu1995image}). For fusion models trained under the high-quality input assumption, we employ a more comprehensive set of reference-based evaluation criteria, including information theory-based metrics—feature mutual information ($FMI$ \cite{haghighat2011non,qu2002information}) and mutual information ($MI$ \cite{qu2002information}); human perception inspired metric—visual information fidelity ($VIF$ \cite{han2013new}); image feature-based metric—gradient-based similarity measurement ($Q_{abf}$ \cite{xydeas2000objective}); as well as two structural similarity-based metrics—Piella's metric ($Q_{w}$ \cite{piella2003new}) and Yang's metric ($Q_{y}$ \cite{yang2008novel}). For all ten metrics, higher values indicate superior fusion performance.

\subsection{Image Fusion under High-Quality Input Assumption}
\subsubsection{Qualitative Analysis}

\begin{figure*}[!t]
\centering
\includegraphics[width=0.75\textwidth]{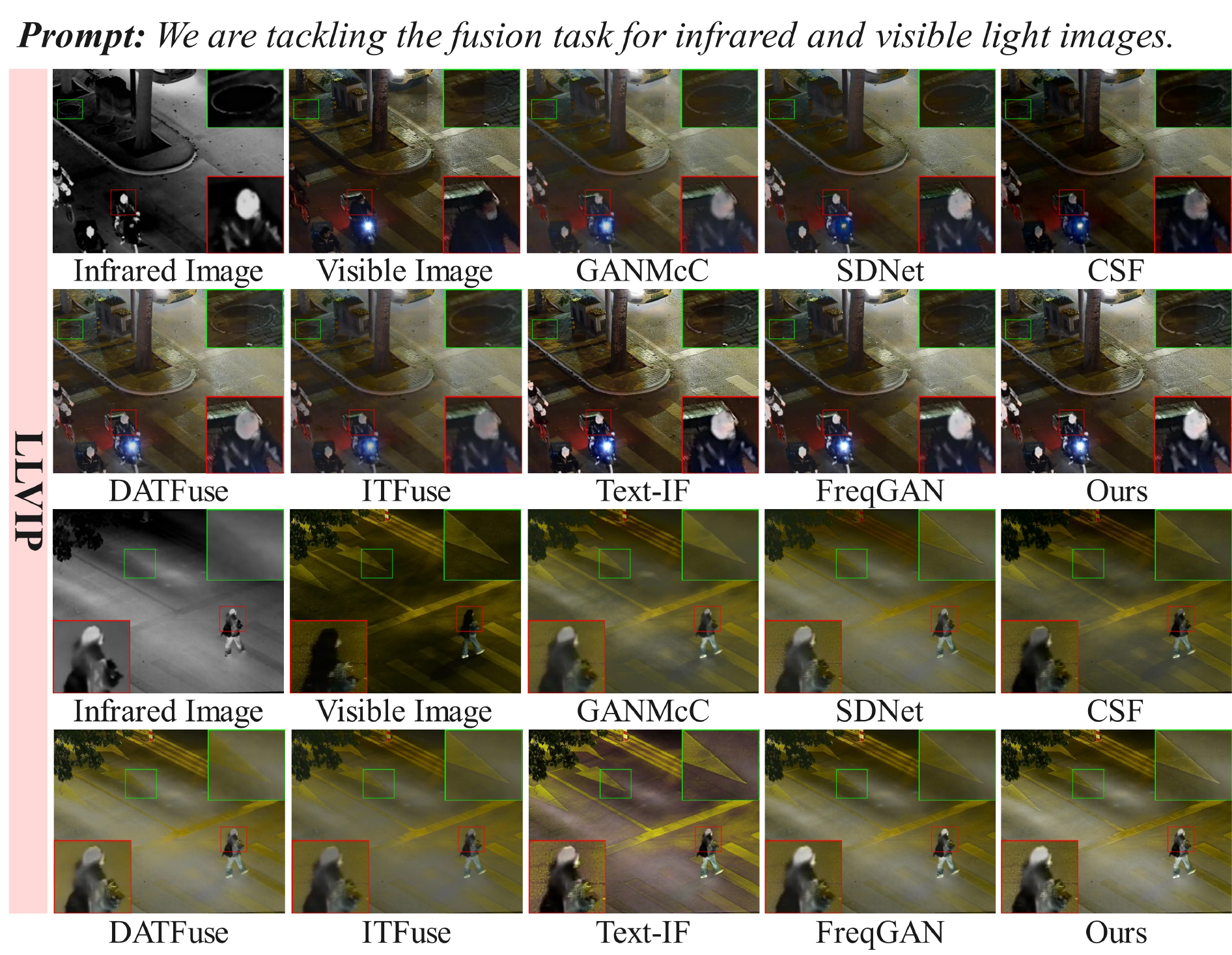}
%图中展示了 GD$^2$Fusion 与其他七种比较图像融合方法在 LLVIP 数据集上进行融合实验的定性对比结果，实验场景基于高质量图像输入的假设设定。图中所示的 Prompt 表示输入到我们网络中的文本提示信息，用于指导融合过程。为了突出关键差异，我们在融合结果中对若干感兴趣区域采用红色与绿色矩形框进行标注，并对其进行了局部放大，从而便于在细节层面开展直观的视觉比较与分析。
\caption{Qualitative comparison results of GD$^2$Fusion against seven comparative image fusion methods on the LLVIP dataset, under the assumption of high-quality input images. The Prompt shown in the figure denotes the textual guidance fed into our network to guide the fusion process. To highlight key differences, several regions of interest are annotated with red and green bounding boxes in the fused results, and subsequently enlarged to facilitate intuitive visual inspection and detailed analysis.}
\label{qualitative_llvip}
\end{figure*}

\begin{table*}[!t]
\centering
%GD$^2$Fusion 与其他七种比较图像融合方法在 MSRS 和 LLVIP 数据集上进行融合实验的定量对比结果，实验场景基于高质量图像输入的假设设定。为了方便比较，我们将每个指标表现最好的三种方法分别用\textcolor{red}{red}，\textcolor{blue}{blue},\textcolor{green}{green}表示，$\uparrow$表示该指标越大所对应的性能越好。
\caption{Quantitative comparison results of GD$^2$Fusion and seven comparative image fusion methods on the MSRS and LLVIP dataset are presented under the assumption of high-quality image inputs. For clarity, the top three performers for each metric are highlighted in \textcolor{red}{red}, \textcolor{blue}{blue}, and \textcolor{green}{green}, respectively, where $\uparrow$ indicates that higher values correspond to better performance.
}
\begin{tabular}{c|ccccccc}
\toprule
\multirow{2}{*}{\textbf{Method}} 
& \multicolumn{6}{c}{\textbf{Infrared and Visible Image Fusion using Fusion prompts(MSRS/LLVIP)}} \\
\cline{2-7}
\noalign{\vskip 1pt} % 可以调节数值（如 1pt, 2pt）
& \textbf{FMI$\uparrow$} & \textbf{MI$\uparrow$} & \textbf{VIF$\uparrow$}& \textbf{Q$_{abf}\uparrow$} & \textbf{Q$_{w}\uparrow$}& \textbf{Q$_{y}\uparrow$} \\
\midrule
GANMcC     
& 0.9365 / 0.9064 & 1.4613 / 1.6884 & 0.6730 / 0.5681 & 0.3439 / 0.2595 & 0.5926 / 0.4825 & 0.6672 / 0.5200 \\
SDNet   
& 0.9354 / 0.9089 & 0.9387 / 1.7956 & 0.5213 / 0.6514 & 0.3514 / \textcolor{green}{0.5328} & 0.6995 / \textcolor{green}{0.7771} & 0.5545 / 0.6340 \\
CSF
& 0.9399 / 0.9067 & 1.1958 / 1.6658 & 0.5436 / 0.6657 & 0.2985 / 0.4198 & 0.6187 / 0.6259 & 0.4369 / 0.5520 \\
DATFuse
& \textcolor{green}{0.9426} / 0.9066 & \textcolor{green}{1.9270} / \textcolor{blue}{2.1911} & \textcolor{green}{0.8853} / 0.7011 & \textcolor{green}{0.5608} / 0.4697 & \textcolor{green}{0.8468} / 0.6835 & \textcolor{green}{0.7705} / \textcolor{green}{0.7242} \\
ITFuse
& 0.9360 / 0.9021 & 1.3660 / 1.7677 & 0.5908 / 0.5658 & 0.2753 / 0.2124 & 0.5061 / 0.3862 & 0.5539 / 0.5290 \\
Text-IF
& \textcolor{blue}{0.9510} / \textcolor{blue}{0.9194} & \textcolor{blue}{2.0586} / 1.9294 & \textcolor{red}{1.0159} / \textcolor{blue}{0.9333} & \textcolor{blue}{0.6088} / \textcolor{blue}{0.6624} & \textcolor{blue}{0.9018} / \textcolor{blue}{0.8877} & \textcolor{blue}{0.8443} / \textcolor{blue}{0.7279} \\
FreqGAN
& 0.9421 / \textcolor{green}{0.9132} & 1.3754 / \textcolor{green}{2.0151} & 0.6426 / \textcolor{green}{0.7170} & 0.4526 / 0.4637 & 0.7860 / 0.6717 & 0.6911 / 0.6937 \\
\textbf{Ours}     
& \textcolor{red}{0.9530} / \textcolor{red}{0.9273} & \textcolor{red}{2.2290} / \textcolor{red}{2.5573} & \textcolor{blue}{0.9927} / \textcolor{red}{0.9881} & \textcolor{red}{0.6383} / \textcolor{red}{0.7351} & \textcolor{red}{0.9167} / \textcolor{red}{0.9206} & \textcolor{red}{0.8543} / \textcolor{red}{0.7761} \\
\bottomrule
\end{tabular}
\label{tab:quantitative_normal}
\end{table*}

\begin{figure*}[!t]
\centering
\includegraphics[width=0.75\textwidth]{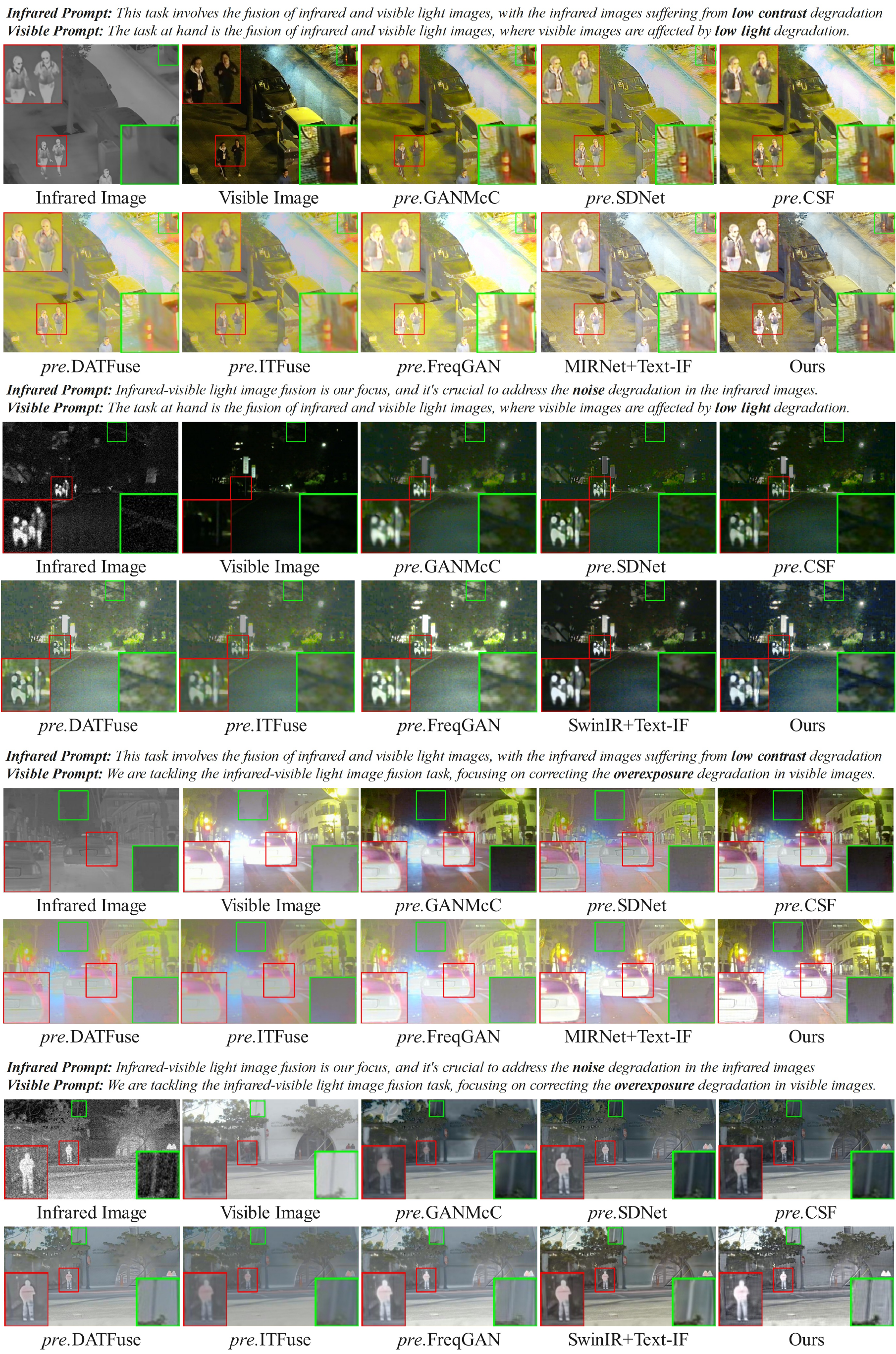}
%GD$^2$Fusion与其他七种图像融合方法在EMS数据集中的退化图像融合的定性比较，源图像退化为可见光的低光/过曝退化及红外的低对比度/噪声退化的组合。\(\textit{pre.}\)代表解决两种源图像退化的图像预增强方法（可见光的亮度分布调整：IAT，红外图像的对比度增强：MIRNet，红外图像的去噪：SwinIR），我们的GD$^2$Fusion所用到的红外/可见光提示文本在图像对比的上方展示。
\caption{Qualitative comparisons of GD$^2$Fusion against seven comparative image fusion methods are conducted on the EMS dataset under degraded image fusion scenarios, where the source degradations consist of low-light/overexposure in visible images and low-contrast/noise in infrared images. $\textit{pre.}$ denotes image pre-enhancement strategies applied to address source degradations. The infrared/visible textual prompts used in our GD$^2$Fusion are displayed above the comparative images. }
\label{qualitative_degrade}
\end{figure*}

\begin{table*}[!t]
\centering
%GD$^2$Fusion与其他七种图像融合方法在EMS数据集中的退化图像融合的定量比较，源图像退化为可见光的低光/过曝退化及红外的低对比度/噪声退化的组合。\(\textit{pre.}\)代表解决两种源图像退化的图像预增强方法（可见光的亮度分布调整：IAT，红外图像的对比度增强：MIRNet，红外图像的去噪：SwinIR），值得注意的是，对于Text-IF，\(\textit{pre.}\)表示使用了红外图像的预增强方法，而可见光图像通过其预训练模型实现退化去除。为了方便比较，我们将每个指标中表现前三的方法分别用\textcolor{red}{red}，\textcolor{blue}{blue},\textcolor{green}{green}表示，$\uparrow$表示该指标越大所对应的性能越好。
\caption{Quantitative comparison of degraded image fusion on the EMS dataset between GD$^2$Fusion and seven comparative fusion methods, where the source images are degraded by combinations of low-light/over-exposure in the visible modality and low-contrast/noise in the infrared modality. $\textit{pre.}$ denotes pre-enhancement strategies designed to address degradations in the two modalities (luminance distribution adjustment for visible images: IAT; contrast enhancement for infrared images: MIRNet; denoising for infrared images: SwinIR). For ease of comparison, the top three methods for each metric are highlighted in \textcolor{red}{red}, \textcolor{blue}{blue}, and \textcolor{green}{green}, respectively, where $\uparrow$ indicates that a higher value corresponds to better performance.}
\begin{tabular}{c|cccc|cccc}
\toprule
\multirow{3}{*}{\textbf{Method}} 
& \multicolumn{8}{c}{\textbf{Infrared and Visible Image Fusion using Degradation-aware fusion prompts}} \\
\cline{2-9}
\noalign{\vskip 2pt} % 可以调节数值（如 1pt, 2pt）
&\multicolumn{4}{c|}{\textbf{Low light \& Low Contrast}} & \multicolumn{4}{c}{\textbf{Overexposure \& Low Contrast}} \\
& \textbf{AG$\uparrow$} & \textbf{EI$\uparrow$} & \textbf{SD$\uparrow$} & \textbf{SF$\uparrow$} & \textbf{AG$\uparrow$} & \textbf{EI$\uparrow$} & \textbf{SD$\uparrow$} & \textbf{SF$\uparrow$} \\
\midrule
\textit{pre.}GANMcC    
& 3.0361 & 32.1297 & 33.9925 & 8.4501 & 3.9382 & 41.5315 & 43.0707 & 10.3083 \\
\textit{pre.}SDNet
& \textcolor{blue}{4.2944} & \textcolor{blue}{44.9073} & 30.1818 & \textcolor{blue}{12.7675} & \textcolor{green}{5.5505} & 57.5023 & 30.2246 & \textcolor{green}{15.0173} \\
\textit{pre.}CSF
& 3.6910 & 39.0452 & 38.9161 & 10.5548 & 5.5136 & \textcolor{green}{58.2206} & \textcolor{blue}{51.6323} & 14.2895 \\
\textit{pre.}DATFuse
& 3.9760 & 41.5518 & 39.9126 & 11.7116 & 2.4688 & 24.8667 & 20.0634 & 6.9313 \\
\textit{pre.}ITFuse
& 2.6916 & 28.8046 & 31.1906 & 7.0680 & 1.7973 & 19.3473 & 21.1845 & 4.5589 \\
\textit{pre.}FreqGAN
& 3.9954 & 42.6283 & \textcolor{red}{51.4058} & 11.6555 & 2.9447 & 30.9695 & 24.1970 & 7.6271 \\
MIRNet + Text-IF
& \textcolor{green}{4.1750} & \textcolor{green}{43.5687} & \textcolor{green}{40.2276} & \textcolor{green}{12.1471} & \textcolor{blue}{5.7382} & \textcolor{blue}{60.0273} & \textcolor{green}{43.2465} & \textcolor{blue}{16.2814} \\
\textbf{Ours}     
& \textcolor{red}{4.5904} & \textcolor{red}{47.6333} & \textcolor{blue}{43.1937} & \textcolor{red}{14.0132} & \textcolor{red}{6.9128} & \textcolor{red}{71.8015} & \textcolor{red}{52.3444} & \textcolor{red}{19.6723} \\
\midrule
& \multicolumn{4}{c|}{\textbf{Low light \& Noise}} & \multicolumn{4}{c}{\textbf{Overexposure \& Noise}}\\
& \textbf{AG$\uparrow$} & \textbf{EI$\uparrow$} & \textbf{SD$\uparrow$} & \textbf{SF$\uparrow$} & \textbf{AG$\uparrow$} & \textbf{EI$\uparrow$} & \textbf{SD$\uparrow$} & \textbf{SF$\uparrow$}\\
\midrule
\textit{pre.}GANMcC    
& 2.8001 & 29.6527 & 32.5217 & 8.1019 & 3.6763 & 39.2594 & 47.3254 & 9.2985 \\
\textit{pre.}SDNet
& 3.5890 & 37.5978 & 27.9205 & 11.5728 & \textcolor{green}{5.5561} & \textcolor{green}{58.2337} & 46.8483 & \textcolor{green}{14.9645} \\
\textit{pre.}CSF
& 3.1236 & 33.2104 & 36.1315 & 9.5639 & 4.8479 & 51.4196 & \textcolor{blue}{51.4207} & 12.7233 \\
\textit{pre.}DATFuse
& \textcolor{blue}{4.2706} & \textcolor{blue}{44.7484} & \textcolor{green}{42.1463} & \textcolor{blue}{12.8840} & 3.9580 & 40.3148 & 34.0890 & 11.8683 \\
\textit{pre.}ITFuse
& 2.6612 & 28.5616 & 31.8886 & 7.1525 & 2.3799 & 25.6738 & 28.8083 & 6.0372 \\
\textit{pre.}FreqGAN
& \textcolor{green}{3.9710} & \textcolor{green}{42.5557} & \textcolor{red}{53.1112} & 12.1019 & 4.0825 & 43.3966 & 44.3393 & 11.3119 \\
Swin-IR + Text-IF
& 3.9242 & 41.1069 & 41.1377 & \textcolor{green}{12.2359} & \textcolor{blue}{5.6621} & \textcolor{blue}{59.5389} & \textcolor{green}{51.2610} & \textcolor{blue}{15.6672} \\
\textbf{Ours}     
& \textcolor{red}{4.5412} & \textcolor{red}{47.3899} & \textcolor{blue}{43.3764} & \textcolor{red}{13.9295} & \textcolor{red}{6.6836} & \textcolor{red}{69.5317} & \textcolor{red}{52.9414} & \textcolor{red}{19.0150}\\
\bottomrule
\end{tabular}
\label{tab:quantitative_degrade}
\end{table*}

%说明每个方面的重要性+对比方法的具体说明
%Fig..\ref{qualitative_msrs} 与 Fig..\ref{qualitative_llvip} 分别给出了 GD$^2$Fusion 与七种典型对比方法在基于高质量图像输入假设下的主观融合效果对比。总体而言，这些对比方法均表现出不同程度的性能局限，可从以下几个方面进行归纳：（1）红外显著目标的继承能力。红外图像中的显著目标信息通常是融合结果的核心表征。然而，除 Text-IF 和 FreqGAN 外，其余方法在不同程度上均存在对红外目标强度信息继承不足的现象，这在 Fig..\ref{qualitative_msrs} 与 Fig..\ref{qualitative_llvip} 的红色矩形框所示区域中表现得尤为明显。（2）红外结构轮廓的补充利用。在低照度场景中，可见光图像的结构特征常难以保持清晰度，此时需依赖红外图像提供边缘与轮廓信息，以帮助区分前景与背景。然而，GANMcC、CSF、ITFuse、 Text-IF 以及 FreqGAN 在该互补关系的建模上均存在不足。如 Fig..\ref{qualitative_msrs} 前两行的绿色矩形框所示，窗户与墙壁之间的边界不清晰；在后两行的绿色矩形框中，屋檐与天空的分界亦出现模糊。（3）可见光细节的保真性。作为融合图像感知质量的重要组成部分，可见光细节若得不到充分保留，将直接削弱融合图像的清晰度与视觉可读性。然而，GANMcC、SDNet、CSF、 ITFuse 以及 FreqGAN 在可见光细节边缘保持方面表现不足。例如 Fig..\ref{qualitative_llvip} 前两行的绿色矩形框所示，地砖边界模糊甚至消失，显著降低了图像的视觉质量。（4）对红外信息的过度依赖。理想的融合结果应在红外与可见光信息之间实现平衡继承，从而发挥互补优势。但如 Fig..\ref{qualitative_llvip} 第四行所示，DATFuse 明显偏向红外信息，导致可见光图像中的结构轮廓几乎完全丢失。（5）色彩失真问题。色彩保真性是保证融合结果视觉自然性与真实感的关键。然而，Text-IF 在部分场景下由于红外强度信息的引入出现明显的颜色偏差，例如 Fig..\ref{qualitative_llvip} 第四行所示区域，融合结果呈现出显著的色彩失真。综上所述，相较于这些对比方法，GD$^2$Fusion 在红外显著目标提取、结构轮廓补充、可见光细节保留、跨模态信息平衡以及色彩保真性等多个维度均展现出更优越的表现，从而生成了信息表达更为全面、视觉质量更为自然友好的融合结果。
Fig.~\ref{qualitative_msrs} and Fig.~\ref{qualitative_llvip} present the qualitative comparison of GD$^2$Fusion against seven comparative fusion methods under the assumption of high-quality image inputs. Overall, these comparative approaches exhibit varying degrees of performance limitations, which can be summarized as follows: (1) \textbf{Inheritance of salient infrared targets}: Salient object information in infrared images typically constitutes the core representation of the fused output. However, except for Text-IF and FreqGAN, other five comparative methods demonstrate insufficient inheritance of infrared target intensity to different extents, as is particularly evident in the regions highlighted by red rectangles in Fig.~\ref{qualitative_msrs} and Fig.~\ref{qualitative_llvip}. (2) \textbf{Supplementary utilization of infrared structural contours}: In low-illumination scenarios, structural features in visible images often lack clarity, thereby necessitating infrared information to provide edge and contour cues for distinguishing foreground from background. Nevertheless, GANMcC, CSF, ITFuse, Text-IF and FreqGAN show weaknesses in modeling such complementarity. As illustrated by the green rectangles in the first two rows of Fig.~\ref{qualitative_msrs}, the boundary between windows and walls becomes indistinct; likewise, in the last two rows, the roof–sky separation is blurred. (3) \textbf{Fidelity of visible details}: As an essential factor of perceived fusion quality, inadequate preservation of visible details directly degrades the sharpness and readability of the fused images. GANMcC, SDNet, CSF, ITFuse and FreqGAN exhibit insufficient preservation of visible edges; for instance, the green rectangles in the first two rows of Fig.~\ref{qualitative_llvip} reveal blurred or even vanished floor tile boundaries, leading to a notable decline in visual quality. (4) \textbf{Over-reliance on infrared information}: An ideal fusion output should achieve a balanced inheritance of infrared and visible information to fully exploit their complementary properties. However, as shown in the fourth row of Fig.~\ref{qualitative_llvip}, DATFuse is clearly biased towards infrared content, resulting in the near-complete loss of structural contours from visible images. (5) \textbf{Color distortion}: Preserving color fidelity is vital for ensuring the naturalness and realism of fused images. Yet Text-IF exhibits noticeable color deviations in certain scenes due to the introduction of infrared intensity information. For example, the highlighted area in the fourth row of Fig.~\ref{qualitative_llvip} demonstrates severe color distortion in the fusion output. In summary, compared with these competing methods, GD$^2$Fusion demonstrates superior performance across multiple dimensions—including salient infrared target extraction, structural contour supplementation, visible detail preservation, balanced cross-modality information integration, and color fidelity—ultimately producing fused results with richer informational representation and more natural, visually coherent quality.

\subsubsection{Quantitative Analysis}
%Tab..~\ref{tab:quantitative_normal}展示了GD$^2$Fusion与七种代表性对比方法在MSRS与LLVIP数据集上，在高质量图像输入假设下的融合结果的定量性能对比。由表中结果可见，除在MSRS数据集的$VIF$指标上略逊于Text-IF外，GD$^2$Fusion在MSRS上的其余五项指标均取得最佳表现，同时在LLVIP数据集的全部六项指标上也显著优于所有对比方法。这一结果充分验证了本方法在信息传递与保持、图像细节与特征保留、结构一致性以及视觉感知质量等多个层面均展现出优越的融合能力，表明所提出的框架能够在不同数据环境下稳定输出具有较高感知与结构质量的融合图像。
Tab.~\ref{tab:quantitative_normal} presents the quantitative performance comparison of GD$^2$Fusion against seven representative comparative methods on the MSRS and LLVIP datasets under the assumption of high-quality image inputs. As shown in the results, GD$^2$Fusion achieves the best performance on five out of six metrics on the MSRS dataset, with only a slight inferiority to Text-IF in terms of the $VIF$ metric, while it consistently outperforms all comparison methods across all six metrics on the LLVIP dataset. These results clearly demonstrate the superior fusion capability of the proposed method in terms of information transfer and preservation, image detail and feature retention, structural consistency, and perceptual quality, thereby indicating that the framework is capable of generating fused images with high perceptual and structural quality in diverse data scenarios.

\begin{figure*}[!t]
\centering
\includegraphics[width=\textwidth]{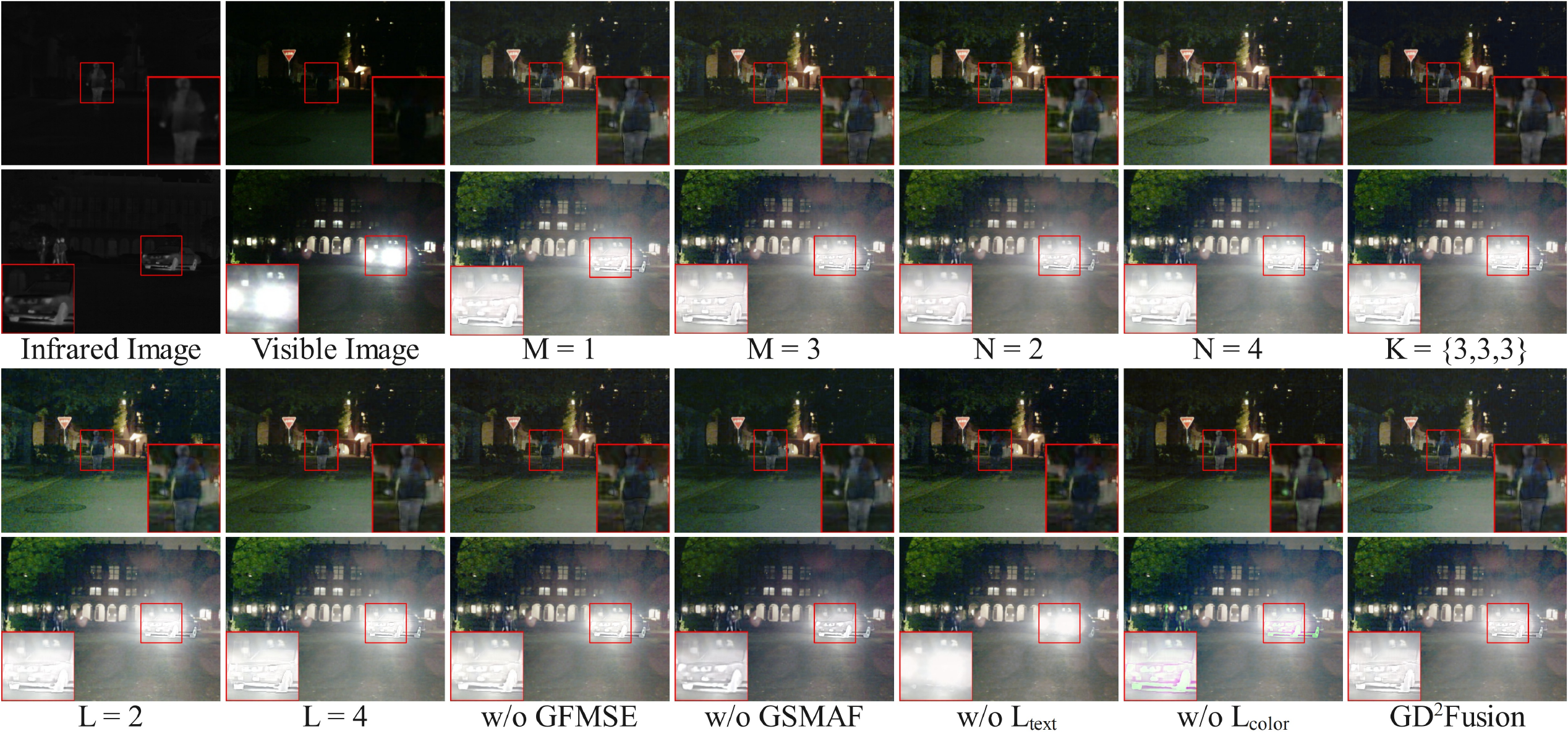}
%GD$^2$Fusion的消融实验的定性分析，其中M和N分别代表GFMSE以及GSMAF中Transformer和卷积块的堆叠数量，K代表GSMAF中多尺度卷积的卷积核尺寸，L代表GFMSE以及GSMAF的堆叠层数。为了方便视觉比较，我们将图中的重点区域用红框进行框选并放大
\caption{Qualitative analysis of the ablation study for GD$^2$Fusion, where $M$ and $N$ denote the number of stacked Transformer and convolutional blocks in GFMSE and GSMAF, respectively; $K$ represents the kernel size of the multi-scale convolutions in GSMAF; and $L$ indicates the number of stacked layers in both GFMSE and GSMAF. For clearer visual comparison, key regions in the images are highlighted with red bounding boxes and magnified.}
\label{qualitative_ablation}
\end{figure*}

\subsection{Image Fusion under Degraded Input Conditions}
%为在“双源均退化”的融合场景下公平验证所提方法的优势，我们对不同类型的对比方法采用一致的评测协议：对于基于高质量输入假设的融合网络，先使用先进的图像预增强算法对退化源图像进行处理，再将增强后的结果送入其融合模型以获得用于比较的输出；对于面向特定退化的融合方法（如 Text-IF），则按照官方提供的预训练模型对其中一类模态的退化进行去除，同时对另一模态执行预增强以满足其对“干净输入”的要求。具体而言，针对可见光图像中的低光与过曝退化，以及红外图像中的低对比度与噪声退化，我们采用 MIRNet 进行红外对比度增强，使用 Swin-IR 去除红外噪声，并通过 IAT 调整可见光图像的亮度分布，从而实现多类型退化的组合优化与在同一评测标准下的公平对比。
To ensure a fair evaluation of the proposed method under the “dual-source degradation” fusion scenario, we adopt a unified evaluation protocol for different types of comparison methods. Specifically, for fusion networks designed under the assumption of high-quality inputs, advanced image pre-enhancement algorithms are first applied to the degraded source images, and the enhanced results are subsequently fed into their fusion models to obtain outputs for comparison. For degradation-oriented fusion methods (e.g., Text-IF), the official pretrained models are employed to remove degradation in one modality, while the other modality undergoes pre-enhancement to satisfy their requirement for “clean inputs.” Concretely, for degradations in visible images such as low-light and overexposure, and in infrared images such as low contrast and noise, we utilize MIRNet \cite{Zamir2022MIRNet} for contrast enhancement of infrared images, Swin-IR \cite{liang2021swinir} for infrared noise removal, and IAT \cite{Cui_2022_IAT} for brightness distribution adjustment in visible images, thereby achieving combined optimization across multiple degradation types and ensuring fair comparisons under a consistent evaluation standard.

\subsubsection{Qualitative Analysis}
%Fig..~\ref{qualitative_degrade} 给出了 GD$^2$Fusion 与七种对比方法在四类退化场景下的定性对比。总体可见，各对比方法在不同情形下均存在一定局限，具体体现在：(1) 红外低对比度 \& 可见光低光： 多数方法生成的融合结果对红外显著目标的强度继承不足，或整体对比度偏低。除 FreqGAN 和 Text-IF 外，其余六种方法对红外强度的保留均不同程度减弱，表现为 Fig..~\ref{qualitative_degrade} 前两行红框中的行人未被充分“高亮”。同时，尽管可见光经预增强已具备合适的强度，SDNet、DATFuse、ITFuse、FreqGAN 与 Text-IF 的结果仍继承了大量红外的强度分布，导致对比度偏低，削弱了暗部细节的可辨性（见前两行绿框）。(2) 红外噪声 \& 可见光低光: 该情景下常见问题包括红外结构边缘模糊、融合颜色偏差以及整体对比度不足。具体而言，GANMcC、ITFuse以及FreqGAN 的融合结果在显著目标的轮廓边缘处出现明显模糊（Fig..~\ref{qualitative_degrade} 第 3–4 行红框）；除 Text-IF 外，其余方法均存在显著的色偏,影响融合图像的视觉感知质量；此外，除 DATFuse 以及 FreqGAN 外，其余六种方法在低光增强后仍呈现低对比度，直接影响到树枝/树叶与天空之间的区分与细节呈现（见第 3–4 行绿框）。(3) 红外低对比度 \& 可见光过曝: 对比方法在过曝区域细节补充与前景/背景对比度关系处理上均表现欠佳。Fig..~\ref{qualitative_degrade} 第 5–6 行红框所示的过曝区域中，DATFuse 与 ITFuse 难以利用增强后的红外边缘进行有效补充，车牌与车灯边缘近乎消失；同时，当两源图像在前景与背景呈现相反对比度关系时，七种方法均未能正确处理，导致分界线模糊（第 5–6 行绿框中树叶与天空几乎不可分）。(4) 红外噪声 \& 可见光过曝: 该情景下，各方法普遍存在显著目标强度继承不足的问题（Fig..~\ref{qualitative_degrade} 第 7–8 行红框）；除 DATFuse 外，其余方法对同一区域内两源图像的相反对比度关系处理仍不理想，表现为电线与树叶边缘呈现不自然的视觉感受及削弱的边缘表达（见第 7–8 行绿框）。综合上述定性结果可见，GD$^2$Fusion 相较于各对比方法在退化场景下更具优势：其在端到端的统一优化框架内同时实现退化感知、退化抑制与图像融合，避免了预处理与融合阶段相互分离所造成的信息损失；并且能够更充分地发挥跨模态互补与双域交互的作用，使红外显著目标信息得以有效继承、可见光细节与结构信息的有效提取 以及 两源信息的有效互补，从而获得信息表征能力更强、视觉质量更优的融合结果。
Fig.~\ref{qualitative_degrade} presents the qualitative comparison between GD$^2$Fusion and seven comparative fusion methods under four categories of degradation scenarios. Overall, it can be observed that each comparative method exhibits certain limitations under different conditions, specifically manifested as follows: (1) Infrared low contrast \& visible low light: Most methods fail to adequately preserve the intensity of salient infrared targets or produce fusion results with insufficient overall contrast. Except for FreqGAN and Text-IF, the other five approaches show varying degrees of attenuation in infrared intensity retention, as evidenced by the pedestrians in the red boxes of the first two rows in Fig.~\ref{qualitative_degrade}, which are not sufficiently “highlighted.” Meanwhile, despite the pre-enhancement of the visible image providing suitable brightness, the results of SDNet, DATFuse, ITFuse, FreqGAN and Text-IF still inherit excessive infrared intensity distributions, resulting in diminished contrast and reduced discernibility of dark details (green boxes in the first two rows). (2) Infrared noise \& visible low light: Common issues in this case include blurred infrared structural edges, color distortion in the fused image, and inadequate contrast. Specifically, the fusion results of GANMcC, ITFuse and FreqGAN exhibit noticeable blurring along the contours of salient objects (red boxes in rows 3–4 of Fig.~\ref{qualitative_degrade}); except for Text-IF, all methods suffer from pronounced color deviation, degrading the perceptual quality of the fused image. Furthermore, with the exception of DATFuse and FreqGAN, the remaining five approaches still produce low-contrast results even after low-light enhancement, directly impairing the distinction and detail presentation between tree branches/leaves and the sky (green boxes in rows 3–4). (3) Infrared low contrast \& visible overexposure: In this scenario, the compared methods perform poorly in recovering overexposed details and in handling the contrast relationship between foreground and background. As shown in the red boxes of rows 5–6 in Fig.~\ref{qualitative_degrade}, DATFuse and ITFuse fail to effectively utilize enhanced infrared edges for detail recovery, causing license plate and headlight boundaries to nearly vanish. Moreover, when the two source images present opposite contrast relationships between foreground and background, none of the seven methods properly resolves the issue, leading to blurred boundaries where tree leaves and sky become indistinguishable (green boxes in rows 5–6). (4) Infrared noise \& visible overexposure: In this setting, all methods generally suffer from inadequate inheritance of salient infrared intensity (red boxes in rows 7–8 of Fig.~\ref{qualitative_degrade}). Except for DATFuse, the other methods still struggle with the inconsistent contrast relationships between the two modalities in the same regions, resulting in unnatural visual impressions and weakened edge representation for structures such as wires and tree leaves (green boxes in rows 7–8). In summary, the above qualitative results demonstrate that GD$^2$Fusion exhibits clear advantages over competing approaches in degraded scenarios. By unifying degradation perception, suppression, and image fusion within an end-to-end optimization framework, it effectively avoids the information loss typically caused by separating preprocessing and fusion stages. Moreover, GD$^2$Fusion fully exploits cross-modality complementarity and dual-domain interactions, enabling robust inheritance of salient infrared targets, effective extraction of visible structural and detail information, and comprehensive cross-source information integration. Consequently, the proposed method achieves stronger representational capacity and superior visual quality in the fused results.

\subsubsection{Quantitative Analysis}
%表~\ref{tab:quantitative_degrade} 列示了在四类退化图像融合情形下，GD$^2$Fusion 与七种对比方法的定量比较结果。具体来看：在低光场景中，尽管 FreqGAN 在 $SD$ 指标上取得了较优表现，但其在该场景的其余评价项上与 GD$^2$Fusion 存在较大差距，且在过曝场景中出现了明显的性能下降。对于其余退化情形，GD$^2$Fusion 在所有考察的指标上均取得了领先地位。综上所述，定量实验结果进一步证明了 GD$^2$Fusion 在处理多种退化条件下的融合问题时，能够更稳定且全面地提升融合质量。
Tab.~\ref{tab:quantitative_degrade}. presents the quantitative comparison results of GD$^2$Fusion against seven representative methods under four categories of degraded image fusion scenarios. Specifically, in low-light conditions, although FreqGAN achieves relatively favorable performance on the $SD$ metric, it exhibits a substantial gap compared with GD$^2$Fusion across the remaining evaluation measures for this scenario, and its performance further degrades markedly in overexposure conditions. For the other degradation cases, GD$^2$Fusion consistently outperforms all competing methods across all assessed metrics. In summary, the quantitative results provide further evidence that GD$^2$Fusion delivers more stable and comprehensive improvements in fusion quality when addressing diverse degradation conditions.

\begin{table*}[!t]
\centering
%GD$^2$Fusion的消融实验的定量分析。我们将每个指标中表现最好三个的分别用红色蓝色绿色标出，$\uparrow$代表值越大所对应的表现越好
\caption{Quantitative analysis of the ablation study for GD$^2$Fusion. The top three results for each metric are highlighted in \textcolor{red}{red}, \textcolor{blue}{blue}, and \textcolor{green}{green}, respectively, where $\uparrow$ indicates that a larger value corresponds to better performance.}
\begin{tabular}{c ccccccc}
\toprule
\multirow{2}{*}{Metric} & \multirow{2}{*}{GD$^2$Fusion} & \multicolumn{5}{c}{\textbf{Details of Core Modules}} \\
\cmidrule(lr){3-7}
 &  & M = 1 & M = 3 & N = 2 & N = 4 & K = \{3,3,3\} \\
\midrule
AG $\uparrow$ & \textcolor{green}{4.5904} & 4.5384 & \textcolor{red}{4.6840} & 4.5778 & \textcolor{blue}{4.6288} & 4.4418 \\
EI $\uparrow$ & \textcolor{green}{47.6334} & 47.0096 & \textcolor{red}{48.4688} & 47.5894 & \textcolor{blue}{47.9404} & 46.0023 \\
SD $\uparrow$ & \textcolor{blue}{43.1937} & 41.9716 & 40.9895 & \textcolor{green}{43.1205} & 42.6308 & 40.5744 \\
SF $\uparrow$ & \textcolor{blue}{14.0132} & 13.8316 & \textcolor{red}{14.1414} & 13.8067 & 13.9194 & 13.5898 \\
\midrule
\multirow{2}{*}{Metric} & \multicolumn{2}{c}{\textbf{Effect of Depth}} & \multicolumn{2}{c}{\textbf{Effect of Core Modules}} & \multicolumn{2}{c}{\textbf{Effect of Loss Component}} \\
\cmidrule(lr){2-3} \cmidrule(lr){4-5} \cmidrule(lr){6-7}
 & L = 2 & L = 4 & w/o GFMSE & w/o GSMAF & w/o $L_{text}$ & w/o $L_{color}$ \\
\midrule
AG $\uparrow$ & 4.5441 & 4.5569 & 4.4151 & 4.0314 & 4.1406 & 4.2927 \\
EI $\uparrow$ & 47.3466 & 47.1921 & 45.8626 & 42.0138 & 43.4135 & 44.4699 \\
SD $\uparrow$ & \textcolor{red}{43.2719} & 42.2742 & 40.7929 & 38.3505 & 42.1468 & 40.8484 \\
SF $\uparrow$ & 13.4049 & \textcolor{green}{13.9988} & 13.1751 & 12.4980 & 12.4551 & 13.4658 \\
\bottomrule
\end{tabular}
\label{tab:quantitative_ablation}
\end{table*}

\subsection{Ablation}
\subsubsection{Qualitative Analysis}
%Fig..~\ref{qualitative_ablation} 给出了对 GD$^2$Fusion 的网络结构与损失项进行消融后的定性对比结果。为便于阐述，我们从四个方面展开分析：（1）核心模块的细节：为验证 GFMSE 与 GSMAF 内部堆叠设置的合理性，分别比较了 Transformer 块堆叠层数 \$M\$、卷积块堆叠层数 \$N\$，以及 GSMAF 多尺度卷积的卷积核尺寸 \$K\$ 对融合效果的影响。结果表明，无论 $M$ 或 $N$ 增减均会削弱对可见光低光区域的增强能力，导致结构边缘变弱（见第一行左侧亭子边缘）。当三条多尺度卷积路径的 $K$ 均设为 3 时，虽能一定程度提升显著目标亮度（第一行框选），但低光退化对融合图像的干扰仍然明显，暗部边缘被进一步弱化，且重复的卷积核尺寸造成细节提取不足（第二行框选）。（2）核心模块堆叠数量的影响：为评估三条路径中核心模块（GFMSE、GSMAF）的堆叠数 $L$，我们对不同 $L$ 进行对比。无论增大或减小 $L$，都会导致源图像细节边缘提取/保留不足，使融合结果出现模糊、次优的细节表征（第四行框选）；同时，\$L\$ 过大时对低光区域的增强效果减弱，暗区边缘强度下降，整体可视清晰度受损（第三行图像整体）。（3）核心模块的作用：去除 GFMSE 会显著削弱融合结果对红外显著目标强度的继承，表明频域特征提取对显著信息保持至关重要（第三行框选）；去除 GSMAF 后，增强后的可见光特征无法被有效整合到最终融合表征中，导致低光区域的细节恢复和亮度改进效果显著下降（第三行整体），说明基于空间域的退化感知、特征聚合与退化信息过滤机制是不可或缺的。（4）损失项的贡献：鉴于 $L_{int}$ 是保障基本强度继承的必要项，消融重点放在纹理与色彩相关的子损失。去除 $L_{text}$ 后，网络难以充分恢复细节与边缘，纹理表现明显退化（第四行框选）；且由于缺失与 $L_{int}$ 的协同约束，对显著目标的提取能力亦有所减弱（第三行框选）。去除 $L_{color}$ 则在继承红外显著强度信息时产生明显色彩错误（第四行框选），表明颜色约束对抑制色偏、提升色彩保真度至关重要。综上，对任一组成要素的改动均会不同程度削弱融合效果，定性结果验证了所提模块与损失设计在整体框架中的协同必要性与合理性。
Fig.~\ref{qualitative_ablation} presents the qualitative comparison results of the ablation study conducted on both the network architecture and loss components of GD$^2$Fusion. For clarity, the analysis is organized into four aspects: 

$\bullet$ Details of the core modules: To validate the rationality of the stacking configurations in GFMSE and GSMAF, we compared the impact of varying the number of Transformer blocks $M$, convolutional blocks $N$, and the kernel size $K$ of the multi-scale convolutions in GSMAF. The results indicate that either increasing or decreasing $M$ or $N$ weakens the enhancement of low-light regions in the visible modality, leading to attenuated structural edges (e.g., the pavilion edge on the left of the first row). When the kernel sizes of the three multi-scale convolutional paths are uniformly set to 3, a certain degree of improvement in salient target brightness can be observed (first row, highlighted region). However, the interference of low-light degradation remains evident, causing further weakening of dark-region edges, and the redundant kernel sizes result in insufficient detail extraction (second row, highlighted region).

$\bullet$ Effect of stacking depth in core modules: To assess the influence of the stacking number $L$ of core modules (GFMSE, GSMAF) across the three paths, comparisons were conducted under different $L$. Both increasing and decreasing $L$ lead to inadequate extraction or preservation of fine edges from the source images, yielding blurred and suboptimal detail representation in the fusion results (fourth row, highlighted region). Moreover, when $L$ is excessively large, the enhancement of low-light regions is weakened, with diminished edge intensity in dark areas, thereby impairing overall visual clarity (third row, overall image). 

$\bullet$ Role of the core modules: Removing GFMSE significantly diminishes the inheritance of salient target intensity from the infrared modality, demonstrating that frequency-domain feature extraction is crucial for preserving salient information (third row, highlighted region). After removing GSMAF, the enhanced visible features cannot be effectively integrated into the final fused representation, resulting in a significant decline in detail restoration and brightness improvement in low-light regions (as shown in the third row overall). This demonstrates that the spatial-domain-based mechanisms of guided degradation perception, feature aggregation, and degradation information filtering are indispensable.

$\bullet$ Contribution of loss components: Since $L_{int}$ is essential for ensuring basic intensity inheritance, the ablation primarily focuses on texture- and color-related sub-losses. Without $L_{text}$, the network fails to sufficiently recover details and edges, with markedly degraded texture performance (fourth row, highlighted region). Furthermore, due to the absence of its collaborative constraint with $L_{int}$, the ability to extract salient targets is also weakened (third row, highlighted region). Removing $L_{color}$, on the other hand, introduces noticeable color distortions when inheriting salient infrared intensity information (fourth row, highlighted region), indicating that color constraints are vital for suppressing color shifts and enhancing color fidelity. 

In summary, any modification to the constituent elements results in varying degrees of degradation in fusion performance, and the qualitative results confirm the necessity and rationality of the proposed module and loss designs within the overall framework.

\subsubsection{Quantitative Analysis}
Tab.~\ref{tab:quantitative_ablation} presents the quantitative ablation results of GD$^2$Fusion from both the network design and loss-term perspectives. First, the sensitivity analysis of intra-module hyperparameters reveals that when the number of stacked Transformer blocks $M$ and convolutional blocks $N$ decreases, all the evaluated metrics deteriorate. Conversely, increasing $M$ yields further improvements on certain metrics, but causes a pronounced decline in $SD$; similarly, enlarging $N$ results in slight gains on $AG$ and $EI$, yet these are insufficient to offset the degradation in $SD$ and $SF$. Adjustments to the multi-scale convolutional kernel size $K$ likewise lead to noticeable drops across all metrics, indicating its critical role in detail restoration and cross-modal interaction—overall, the micro-level configuration of the core modules reflects a trade-off between performance and training complexity. Regarding the stacking depth $L$, the results show that apart from a marginal increase in $SD$ at $L=2$, all other metrics degrade to varying extents, suggesting that the chosen number of stacked layers approximates an optimal balance between performance and complexity. Further structural ablation results indicate that when either the GFMSE or GSMAF module is removed, all four evaluated metrics exhibit varying degrees of performance degradation. This finding clearly validates the importance of both components within the network: GFMSE and GSMAF function at different stages of the fusion process, where prompt-guided perception of degradation patterns is achieved, and complementary as well as synergistic feature interactions between the frequency and spatial domains are facilitated to enhance cross-modality image fusion performance. Finally, eliminating any of the sub-loss terms results in consistent metric declines, underscoring the indispensable regularizing role of texture and color constraints within the training objective. In summary, the quantitative ablation results consistently validate the necessity and effectiveness of the proposed modular design and loss composition.

\subsection{Downstream Application}
\begin{figure*}[!t]
\centering
\includegraphics[width=0.75\textwidth]{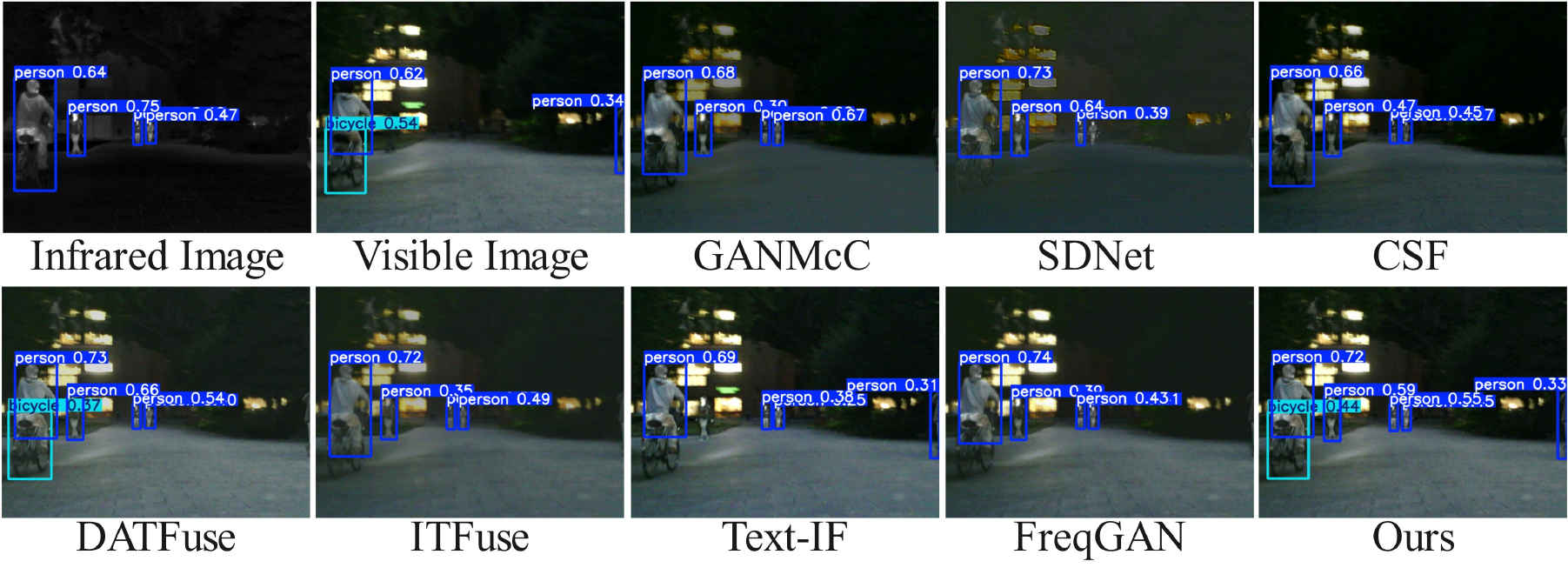}
%GD$^2$Fusion与其他七种图像融合方法在MSRS数据集上进行目标检测任务的定性比较
\caption{Qualitative comparison of GD$^2$Fusion and seven comparative image fusion methods on the MSRS dataset for the object detection task.}
\label{qualitative_detection}
\end{figure*}

\begin{table}[!t]
\centering
%GD$^2$Fusion与其他七种图像融合方法在MSRS数据集上进行目标检测任务的定量比较
\caption{Quantitative comparison of GD$^2$Fusion and seven comparative image fusion methods on the MSRS dataset for the object detection task. The best-performing results for each metric are highlighted in \textcolor{red}{red}.}
\begin{tabular}{cccc}
\hline
\textbf{} & \textbf{mAP@0.85} &  \textbf{mAP@0.90} & \textbf{mAP@[0.5,0.95]} \\
\hline
Infrared Image       & 0.420 &  0.262 & 0.588 \\
Visible Image        & 0.355 &  0.225 & 0.553 \\
GANMcC  & 0.534 &  0.274 & 0.656 \\
SDNet       & 0.487 &  0.232 & 0.606 \\
CSF        & 0.508 &  0.253 & 0.641 \\
DATFuse    & 0.497 & 0.250 & 0.645 \\
ITFuse   & 0.493 & 0.225 & 0.640 \\
Text-IF & 0.491 & 0.324 & 0.644 \\
FreqGAN & 0.487 & 0.237 & 0.629 \\
Ours      & \textcolor{red}{0.588} &  \textcolor{red}{0.336} & \textcolor{red}{0.659} \\
\hline
\end{tabular}
\label{tab:quantitative_detection}
\end{table}
%为验证 GD$^2$Fusion 在下游高层视觉任务中的适用性与优势，我们选取 MSRS 数据集中的 80 对带有标注的图像对进行融合实验，并将所得融合结果进一步输入至 YOLOv5 检测网络中进行评估。随后，我们对检测结果从定性与定量两个层面进行了系统比较，以全面展示所提出方法在提升目标检测性能方面的有效性与潜在应用价值。
To validate the applicability and advantages of GD$^2$Fusion in downstream high-level vision tasks, we conducted fusion experiments on 80 annotated image pairs from the MSRS dataset, and subsequently fed the fused results into the YOLOv5 detection network for evaluation. The detection outcomes were then systematically compared from both qualitative and quantitative perspectives, providing a comprehensive demonstration of the proposed method’s effectiveness in enhancing object detection performance as well as its potential application value.
\subsubsection{Qualitative Analysis}
%Fig.~\ref{Qualitative_Detection} 展示了在目标检测任务中的定性对比结果。与单模态输入相比，采用 GD$^2$Fusion 所生成的融合图像能够有效缓解由单一成像模态固有限制所导致的检测缺失问题。例如，在红外图像中缺乏明显热辐射的自行车，或在低光照环境下可见光图像中难以识别的人体目标，均能够在融合结果中得到更为准确的检测。此外，相较于其他融合方法所产生的结果，GD$^2$Fusion 显示出在多类目标的检测中具有更全面且更稳定的性能表现，能够有效减少漏检现象，如左侧近处的自行车、中远距离的人体，以及右侧存在部分缺失的目标。
Fig.~\ref{qualitative_detection} presents the qualitative comparison results in the context of object detection. Compared with single-modality inputs, the fused images generated by GD$^2$Fusion effectively alleviate detection failures caused by the inherent limitations of individual imaging modalities. For instance, bicycles lacking clear thermal radiation in infrared images or human targets that are difficult to recognize in low-light visible images can be more accurately detected in the fused results. Moreover, relative to the outputs of other fusion methods, GD$^2$Fusion demonstrates more comprehensive and stable performance in multi-class object detection, effectively reducing missed detections, such as nearby bicycles on the left, human figures at medium-to-long distances, and partially occluded targets on the right.

\subsubsection{Quantitative Analysis}
%Tab.~\ref{tab:quantitative_detection} 给出了在目标检测任务上的定量对比结果。可以观察到，在不同的 intersection over union (IoU) 阈值下，GD$^2$Fusion 始终在 mean average precision (mAP) 指标上优于单模态图像以及其他对比融合方法。这一结果表明，GD$^2$Fusion 在下游检测任务中不仅具备更强的目标表征能力，而且能够在不同检测精度要求下保持稳定的优势，从而有力验证了其在目标检测应用中的优越性能。
Tab.~\ref{tab:quantitative_detection} reports the quantitative comparison results on the object detection task. It can be observed that across different intersection over union (IoU) thresholds, GD$^2$Fusion consistently outperforms both single-modality images and other comparative fusion methods in terms of mean average precision (mAP). This outcome indicates that GD$^2$Fusion not only possesses stronger target representation capability in downstream detection tasks but also maintains stable advantages under varying detection accuracy requirements, thereby providing compelling evidence of its superior performance in object detection applications.
% !TEX root = ../main.tex
% \bibliography{../reference.bib}

% \section{Analysis} \label{sec:analysis}
% !TEX root = ../main.tex

\section{Conclusions}
%本文提出了一种名为 \textbf{G}uided \textbf{D}ual-\textbf{D}omain Fusion (GD$^2$Fusion) 的退化感知的图像融合框架，旨在解决现有方法在“双源均退化”场景下频繁依赖人工预增强且融合性能受预增强—融合分离级联策略制约的问题。相较于传统做法，GD$^2$Fusion 在端到端统一优化下将退化感知、抑制与融合协同建模，并系统融入频域信息以增强对细节与结构的表征能力。框架的两个核心模块分别为：频域导向的 \textbf{G}uided \textbf{F}requency \textbf{M}odal-\textbf{S}pecific \textbf{E}xtraction (GFMSE)，该模块借助模态特定的文本提示在频域对单模态退化进行感知与抑制，并判别性地提取有用频带特征；以及空间域的 \textbf{G}uided \textbf{S}patial \textbf{M}odal-\textbf{A}ggregated \textbf{F}usion (GSMAF)，该模块通过聚合模态提示感知并过滤跨模态退化信息，同时在空间域自适应整合多源表示以增强模态互补性与结构一致性。大量定性与定量对比实验表明，GD$^2$Fusion 在多种退化场景下均能显著提升融合图像的细节保真、显著目标继承与视觉质量，验证了所提出双域协同策略与模块设计的有效性与实用价值。
This paper proposes a degradation-aware image fusion framework, termed \textbf{G}uided \textbf{D}ual-\textbf{D}omain Fusion (GD$^2$Fusion), designed to address the limitations of existing approaches in “both-source degraded” scenarios where methods commonly rely on manual pre-enhancement and suffer performance degradation due to a decoupled pre-enhancement - fusion cascade. Unlike conventional pipelines, GD$^2$Fusion jointly models degradation perception, suppression, and fusion under an end-to-end unified optimization and systematically incorporates frequency-domain information to strengthen the representation of fine details and structural cues. The framework comprises two core modules: the frequency-oriented Guided Frequency Modality-Specific Extraction (GFMSE), which leverages modality-specific textual prompts to perceive and suppress unimodal degradations in the frequency domain while discriminatively extracting informative sub-band features; and the spatial-domain Guided Spatial Modality-Aggregated Fusion (GSMAF), which aggregates modality prompts to sense and filter cross-modality degradations and adaptively integrates multi-source spatial representations to enhance modality complementarity and structural consistency. Extensive qualitative and quantitative experiments demonstrate that GD$^2$Fusion substantially improves detail fidelity, salient-target inheritance, and perceptual quality across diverse degradation scenarios, validating the effectiveness and practical value of the proposed dual-domain collaborative strategy and module design.

\bibliographystyle{IEEEtran}
\bibliography{./reference.bib}

\end{document}